\documentclass[pdflatex,sn-nature]{sn-jnl}

\usepackage[switch]{lineno} 

\usepackage{setspace}
\usepackage{graphicx}%
\usepackage{multirow}%
\usepackage{amsmath,amssymb,amsfonts}%
\usepackage{amsthm}%
\usepackage{mathrsfs}%
\usepackage[title]{appendix}%
\usepackage{xcolor}%
\usepackage{textcomp}%
\usepackage{manyfoot}%
\usepackage{booktabs}%
\usepackage{algorithm}%
\usepackage{float} 
\usepackage{placeins}  

\usepackage{algorithmicx}%
\usepackage{algpseudocode}%
\usepackage{listings}%

\theoremstyle{thmstyleone}%
\theoremstyle{thmstyletwo}%

\theoremstyle{thmstylethree}%

\begin{document}


\title[Article Title]{Morphologically Intelligent Perturbation Prediction with \textsc{Form}}

\author[1,2]{\fnm{Reed} \sur{Naidoo}}
\author[2]{\fnm{Matt} \sur{De Vries}}
\author[1]{\fnm{Olga} \sur{Fourkioti}}
\author[2]{\fnm{Vicky} \sur{Bousgouni}}
\author[1]{\fnm{Mar} \sur{Arias-Garcia}}
\author[1]{\fnm{Maria} \sur{Portillo-Malumbres}}
\author*[1, 2]{\fnm{Chris} \sur{Bakal}}

\affil[1]{ \orgname{The Institute of Cancer Research}, \city{London}, \country{United Kingdom}}
\affil[2]{\orgname{Sentinal4D}, \city{London}, \country{United Kingdom}}

\abstract{Understanding how cells respond to external stimuli is a central challenge in biomedical research and drug development. Current computational frameworks for modelling cellular responses remain restricted to two-dimensional representations, limiting their capacity to capture the complexity of cell morphology under perturbation. This dimensional constraint poses a critical bottleneck for the development of accurate virtual cell models. Here, we present \textsc{Form}, a machine learning framework for predicting perturbation-induced changes in three-dimensional cellular structure. \textsc{Form} consists of two components: a morphology encoder, trained end-to-end via a novel multi-channel VQGAN to learn compact 3D representations of cell shape, and a diffusion-based perturbation trajectory module that captures how morphology evolves across perturbation conditions. Trained on a large-scale dataset of over 65,000 multi-fluorescence 3D cell volumes spanning diverse chemical and genetic perturbations, \textsc{Form} supports both unconditional morphology synthesis and conditional simulation of perturbed cell states. Beyond generation, \textsc{Form} can predict downstream signalling activity, simulate combinatorial perturbation effects, and model morphodynamic transitions between states of unseen perturbations. To evaluate performance, we introduce \textsc{MorphoEval}, a benchmarking suite that quantifies perturbation-induced morphological changes in structural, statistical, and biological dimensions. Together, \textsc{Form} and \textsc{MorphoEval} work toward the realisation of the 3D virtual cell by linking morphology, perturbation, and function through high-resolution predictive simulation.}

\maketitle


\newpage

\section{Introduction}

The shape of a cell encodes a wealth of information about its identity, internal state, and functional capacity. Morphological features can reflect cytoskeletal organisation, signalling activity, and cell fate decisions --- and in many cases, offer early indicators of disease progression or therapeutic response \cite{doi:10.1126/science.1140324,doi:10.1126/science.aba2894}. When used in tandem with genetic screening, cell morphology is a powerful means by which to identify genes with diverse roles \citep{Verde1995-nh}. As such, cell shape is more than a descriptive attribute; it is a powerful, interpretable readout that can be leveraged to understand and predict biological behaviour. 
\\
As the field matures beyond retrospective classification and profiling, the vision of the virtual cell is rapidly taking shape: a model that can simulate, explain and hypothesise how cells functionally respond to unseen perturbations under novel conditions \citep{noutahi2025virtualcellspredictexplain}. Generative approaches have emerged in this space, aiming to predict or simulate morphological changes under specific treatments \citep{Palma2025,Bou_PhenDiff_MICCAI2024, Lamiable2023,Cui2024,Adduri2025,zhang2025cellfluxsimulatingcellularmorphology}. However, despite the growing number of simulation-based models, there has been limited progress in improving predictive accuracy or deepening the biological interpretability of perturbation effects. 
\\
To gain a more holistic view of morphology, advances in high-throughput microscopy are increasingly moving from flat, two-dimensional (2D) projections to three-dimensional (3D) imaging, enabling the quantification of cell structure at subcellular resolution \cite{Mertz:19,Fischer2011-ql}. These 3D datasets capture rich phenotypic heterogeneity across perturbations and open new opportunities for characterising cellular responses in greater detail. At the same time, the scale and complexity of such data demand new computational approaches for extracting biologically meaningful patterns and linking them to molecular mechanisms of action, disease states, or therapeutic outcomes. Accordingly, recent advances in deep learning have shown that biologically meaningful features can be extracted directly from 3D cell images. Supervised approaches combining geometric deep learning with attention-based multiple-instance learning have demonstrated that morphological embeddings are not only predictive of treatment identity but also informative of downstream signalling responses, thereby establishing a link between cell form and function \citep{DeVries2025}. 
\\
However, most existing perturbation prediction frameworks are built on 2D microscopy data, which fundamentally limits their capacity to holistically study morphological changes in response to perturbation. 2D projections flatten complex 3D structures, often obscuring key spatial features, such as membrane protrusions and organelle localisation, that are critical for understanding how perturbations affect cell state. This loss of spatial information constrains a model’s ability to learn accurate and generalisable representations of phenotypic change, and lose further predictive power as these embeddings are utilised in downstream simulative settings \citep{DeVries2025}. With 3D imaging now increasingly accessible through high-throughput platforms, there is a growing need for virtual cell models that operate natively in three dimensions, capturing the full structural detail of modern microscopy to enable richer embeddings and more sensitive simulations of subtle perturbation-induced morphological change.  
\\
In addition, most generative models simulate perturbation effects by learning conditionally supervised mappings from untreated to treated states. These include architectures such as conditional autoencoders and generative adversarial networks (GANs) \citep{Bou_PhenDiff_MICCAI2024,Lamiable2023,Adduri2025,Palma2025}, transformational mapping of shared covariates between perturbation distributions \citep{Roohani2024}, and optimal transport-based methods \citep{Bunne2024}; all of which typically frame perturbation as a deterministic or cost-minimising transformation between distributions. While these approaches have shown success when perturbation signals are strong and coherent, they often struggle in settings where biological heterogeneity dominates \citep{wu2025perturbenchbenchmarkingmachinelearning}, such as in the presence of cell cycle variation, lineage bias, or context-dependent effects. More flexible frameworks like flow matching \citep{zhang2025cellfluxsimulatingcellularmorphology} aim to overcome some of these limitations by learning continuous velocity fields between conditions, but still implicitly rely on the assumption that a meaningful trajectory exists across treatment states. As a result, current models often fall short in capturing the full spectrum of phenotypic responses, particularly when those responses are stochastic or non-aligned with simple geometric transformations.
\\
To address these limitations, we introduce \textsc{Form}, a virtual cell model that simulates how 3D cellular morphology and function respond to perturbations. \textsc{Form} consists of two core components: (1) a morphology encoder trained via a multi-channel vector-quantised GAN (VQGAN) to learn compact, high-resolution 3D shape representations, and (2) a diffusion-based perturbation trajectory module that simulates how morphology transitions across treatment conditions. Unlike previous frameworks that directly learn unperturbed-perturbed transformations, \textsc{Form} adopts a distribution-centric view, modelling each perturbation as a distinct morphological landscape and enabling transitions to emerge through probabilistic inference rather than deterministic, supervised mapping. This allows \textsc{Form} to capture the stochastic and heterogeneous nature of real perturbation responses while operating natively in three dimensions.
\\
To support a rigorous evaluation of generated morphologies, we also introduce \textsc{MorphoEval}, an open-source benchmarking suite designed to quantify the biological fidelity of perturbation-induced shape changes. \textsc{MorphoEval} integrates structural, statistical, and functional metrics, including shape-based distances, distributional shifts, and downstream signalling predictions, to assess whether generated cells are realistic and biologically meaningful. Together, \textsc{Form} and \textsc{MorphoEval} represent a step towards realising the virtual 3D cell: a predictive, generative model capable of simulating phenotypic responses to perturbation at subcellular resolution.

\section{Results}

\subsection{\textsc{FORM} is a 3D virtual cell toolkit}

\begin{figure}[h!]
    \centering
    \includegraphics[width=\linewidth]{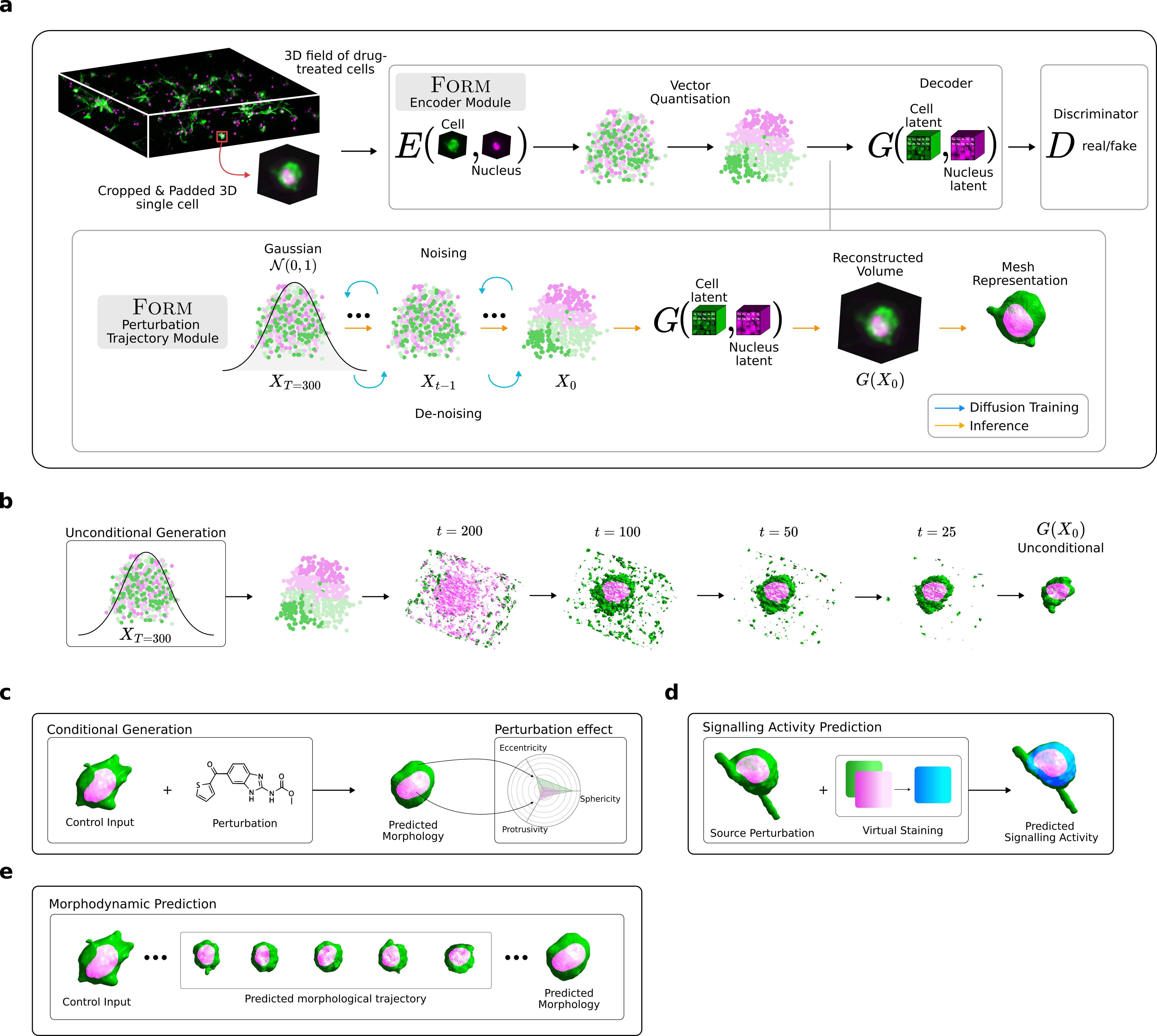}
    \caption{\textbf{Overview of FORM.}
    \textbf{a)} Single-cell 3D volumes are processed through the \textsc{Form} Encoder, and the resulting embeddings are used to train the perturbation trajectory module.
    \textbf{b)} The perturbation trajectory module samples from stochastic noise to generate morphologies under a specified perturbation condition.
    \textbf{c)} Conditioning on a control input, the model generates the corresponding post-treatment morphology and quantifies morphological changes relative to the control.
    \textbf{d)} Predicted morphologies can be further used to simulate intracellular signalling activity directly from structure.
    \textbf{e)} \textsc{Form} also supports modelling of morphodynamic changes, enabling prediction of morphological evolution between perturbation transition states.}
    \label{figure1}
\end{figure}

\textsc{Form} is a two-stage framework that enables the structured representation and drug-perturbed generation of cellular morphologies. The framework consists of two core components: 1) a \textsc{Form} Encoder, a vector-quantised vector adversarial network (VQGAN) \citep{esser2021tamingtransformershighresolutionimage} that learns compact 3D representations of cytoplasmic and nuclear shape, and 2) a \textsc{Form} Trajectory Perturbation Module, a latent multichannel diffusion model \citep{ddpm} that predicts morphological trajectories under perturbation (Figure \ref{figure1}a). 
\\
The first stage of \textsc{Form} trains a VQGAN for each drug perturbation, encoding cellular morphology into learned latent embedding representations (Figure~\ref{figure1}a). The VQGAN follows an encoder–decoder architecture with a vector quantisation step that discretises the latent space into a finite codebook of morphological tokens, ensuring that structural features are represented in a compact and biologically meaningful manner. To further improve reconstruction fidelity and realism, a discriminator is jointly trained in an adversarial fashion, encouraging the decoded volumes to preserve fine-grained morphological detail and phenotypic variability characteristic of real cells.
\\
Although VQGANs and autoencoders have been introduced in high-resolution 3D medical imaging domains \citep{khader2023denoising,Wang2025}, existing approaches treat morphology as a single-channel entity, failing to capture the interdependent relationships between different cellular structures. However, in biological structures (often represented in different colour channels of a microscopy image), the morphological coherence between structures, such as the cytoplasm and the nucleus, is critical for accurate synthesis. Lack of inter-channel and intra-channel consistency in synthetic biological structures can lead to erroneous conclusions, affecting both diagnostic accuracy and treatment evaluation. 
\\
To address this, we introduce a library of independent codebooks, where separate codebooks learn morphological prototypes for the cytoplasm and nucleus channels. During quantisation, each channel is mapped to its closest entry in the corresponding codebook, effectively replacing continuous embeddings with structured, quantised prototypes. 
\\
Following the structured encoding and quantisation of cellular morphology into a discrete latent space, the trained VQGAN is fixed and subsequently used as a morphological tokeniser for all downstream modelling. In the next stage, \textsc{Form} introduces a latent UNet-based \citep{unet,3dunet} denoising diffusion probabilistic model (DDPM) \citep{ddpm,ddim,improvedddim} to enable perturbation-conditioned cell generation. The diffusion model learns to generate latent cellular representations by progressively refining a sampled noise vector into a structured morphological state. The diffusion-based approach allows for controlled sampling from a learned distribution, ensuring that \textsc{Form} captures the heterogeneous morphological responses to drug perturbations. 
\\
The denoised latent is then passed through the pre-trained VQGAN decoder, reconstructing a high-resolution 3D cellular structure that preserves both morphological detail and perturbation specificity. Although the core architecture of \textsc{Form} remains consistent across experiments, the implementation of denoising and decoding at inference directly shapes the trajectory of the generated samples, influencing the diversity, alignment and interpretability of the resulting morphologies. In the sections to follow, we explore how these generative pathways can be configured to synthesise new samples (Figure \ref{figure1}b,c), predict morphological transitions \ref{figure1}e), generate signalling activity \textbf{\ref{figure1}d)}, and model cell relationships across perturbation space. For a detailed description of the training details of \textsc{Form}, please refer to the Online Methods section. 
\\
\subsection{FORM Encodes Multichannel 3D Cellular Morphology for Predicting Biological Relationships}\label{encode}

The capacity of \textsc{Form} to simulate accurate morphological aberrations in response to perturbation is based on the quality and biological precision of its encoded latent representations. Accordingly, the \textsc{Form} encoder and channel-specific codebooks are trained to produce structured morphological embeddings that preserve biologically meaningful variation for downstream morphological analysis. To evaluate the degree to which \textsc{Form} learns biologically meaningful representations, we trained \textsc{Form} Encoder on a dataset of over $65,000$ WM266-4 melanoma cells embedded in collagen matrices (Figure~\ref{figure1}a) and treated with clinically relevant chemical perturbations targeting cytoskeletal and signalling pathways. This training dataset included inhibitors of MEK (binimetinib), myosin-II (blebbistatin), ROCK (H1152), FAK (PF228), CDK4/6 (palbociclib), and microtubules (nocodazole), allowing the model to capture diverse morphological responses to well-characterised drug perturbations. We then applied the pretrained encoder to a distinct dataset of over $35,000$ WM266-4 cells subjected to RNA interference (RNAi), targeting 167 genes across the Rho GTPase signalling axis, including RhoGEFs, RhoGAPs, and Rho family GTPases \citep{DeVries2025,Bousgouni2022-pn}. 
\\
\begin{table}[h]
\begin{tabular}{lcc}
\toprule
Dataset & \textsc{Form} & OpenPhenom \citep{kraus2024masked} \\
\midrule
CORUM \citep{corum} & 0.556 & 0.333 \\
HuMAP \citep{humap} & 0.200 & 0.133 \\
Reactome \citep{reactome} & 0.154 & 0.108 \\
SIGNOR \citep{signor} & 0.177 & 0.106 \\
StringDB \citep{stringdb} & 0.233 & 0.144 \\
\bottomrule
\end{tabular}
\caption{Recall (where higher is better) of known relationships in the top and bottom 5\% of cosine similarities, across methods evaluated on the RNAi dataset. For each dataset, the best-performing normalisation strategy (Typical Variance Normalisation or Centre-Scale) was selected.}
\label{recall}
\end{table}

For each 3D volume, the cytoplasm and nucleus channels were encoded separately, with their respective embeddings concatenated to form a single characteristic vector per cellular volume. These vectors were aggregated per perturbation and normalised to the corresponding DMSO-treated controls within each experimental plate, following the EFAAR (Embedding, Filtering, Aligning, Aggregating, Relating) benchmarking protocol \citep{efaar}. This allowed us to evaluate the degree to which the learnt feature space captured biologically meaningful variation. We computed pairwise cosine similarity scores between aggregated perturbation-level embeddings. Perturbation pairs from the top and bottom 5\% of this similarity distribution were compared to known gene and protein-level interactions curated from CORUM \citep{corum}, huMAP \citep{humap}, Reactome \citep{reactome}, SIGNOR \citep{signor}, and StringDB \citep{stringdb}. To this end, we benchmarked \textsc{Form} against OpenPhenom \citep{kraus2024masked}, an open-source masked autoencoder trained on over 93 million 2D microscopy images for morphological profiling. \textsc{Form} achieved higher recall scores across all four biological reference databases, demonstrating the value of structured, quantised 3D embeddings for uncovering perturbation-induced phenotypic relationships. These results are provided in Table~\ref{recall}. 
\\
\subsection{Unconditional Generation with FORM} \label{unconditional}

\begin{figure}[h!]
    \centering
    \includegraphics[width=\linewidth]{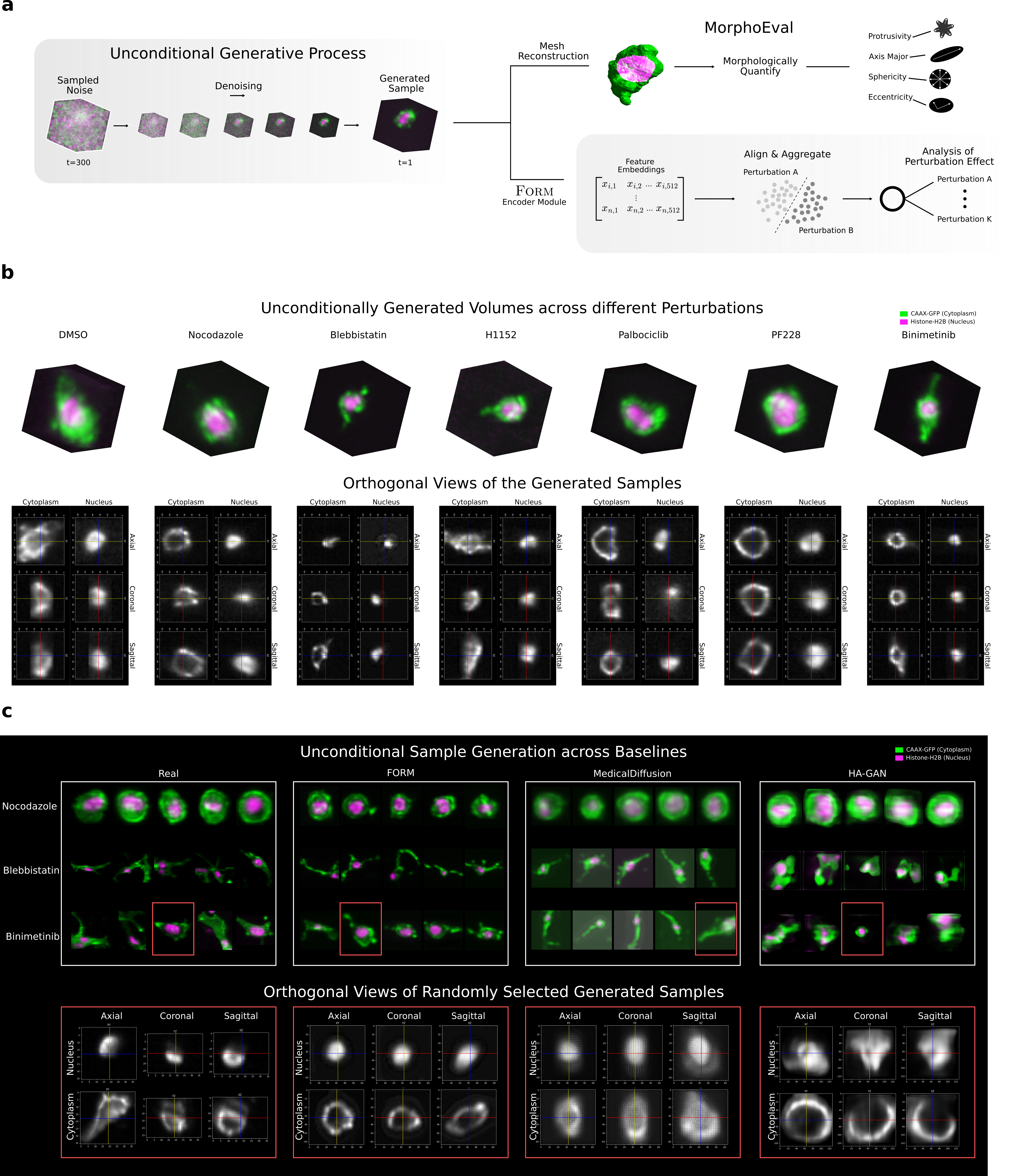}
    \caption{\textbf{Unconditional generative synthesis with FORM.}  
    \textbf{a)} Workflow for analysing unconditional samples. Each generated volume is converted into a mesh for the extraction of morphological descriptors and simultaneously passed through the \textsc{Form} Encoder to obtain feature embeddings for downstream performance evaluation.  
    \textbf{b)} Representative 3D volumes generated under different perturbation settings, with corresponding orthogonal views (axial, coronal, sagittal).  
    \textbf{c)} Comparison of \textsc{Form}-generated samples with state-of-the-art baselines (HA-GAN and MedicalDiffusion) using maximum intensity projections across three representative perturbations: nocodazole, blebbistatin, and binimetinib.}  
    \label{figure3}
\end{figure}

Before \textsc{Form} can be evaluated for its ability to infer morphological trajectories between perturbation-induced cellular states, it must first demonstrate that it faithfully simulates the morphological variability \textit{within} a single perturbation class. In our context, this means demonstrating that \textsc{Form} can generate high-fidelity samples that faithfully reflect the natural morphological heterogeneity observed among cells treated with the same perturbation.
\\
Unlike models trained to predict transitions between perturbation states, \textsc{Form} is trained to capture the full morphological distribution associated with each perturbation. Rather than learning explicit mappings, it models the intra-class variability that arises under a single treatment. Assessing performance in the \textit{unconditional} setting thus provides a direct test of how \textsc{Form} has internalised biologically meaningful intra-class variability.
\\
To evaluate unconditional generative performance, we compared \textsc{Form} with two state-of-the-art 3D medical imaging generative models, MedicalDiffusion \citep{khader2023denoising} and Hierarchical Amortised GAN (HA-GAN) \citep{hagan}, across a subset of perturbation conditions: nocodazole, blebbistatin, and binimetinib. This subset was selected based on prior evidence of their pronounced and visually distinct morphological effects relative to controls \citep{DeVries2025}. All models were trained using the same datasets, and for each perturbation, we synthesised $1,000$ samples and sampled an equal number of real cells for a fair evaluation. All volumes were resized and zero-padded to a standardised shape of ${64}^3$, and the HA-GAN architecture was adjusted accordingly to accommodate this input size.
\\
We first assessed distributional alignment using the Fréchet Inception Distance ($FID$) \citep{fid} and $F1$ score \citep{kynkaanniemi2019improved}, which jointly reflect the fidelity and precision of the generated samples. To further evaluate the diversity of the generated morphologies, we included coverage \citep{naeem2020reliable} as a metric of intraclass heterogeneity. These metrics were applied to features extracted from generated samples using the \textsc{Form} Encoder, as described in Section \ref{encode}. Across the three metrics, \textsc{Form} performed favourably compared to existing baselines, achieving the highest $F1$ score and coverage while maintaining the highest $FID^{-1}$. As shown in Table~\ref{table1}, \textsc{Form} consistently outperforms HA-GAN and MedicalDiffusion on both realism and diversity metrics, with an average improvement of approximately $41\%$. These results suggest that \textsc{Form} better captures the range of phenotypic variation induced by treatment, producing samples that more closely align with real population-level distributions and realism.
\\
\begin{table}[h]
\centering
\begin{tabular}{cccc}
\toprule
Method & $FID^{-1}$ (↑) & F1 Score (↑) & Coverage (↑) \\
\midrule
\textbf{FORM} & $\mathbf{0.822\,(\pm\, 0.183)}$ & $\mathbf{0.57\,(\pm\, 0.097)}$ & $\mathbf{0.741\,(\pm\, 0.112)}$ \\
HA-GAN \citep{hagan} & $0.009\,(\pm\, 0.012)$ & $0.186\,(\pm\, 0.041)$ & $0.651\,(\pm\, 0.128)$ \\
MedicalDiffusion \citep{khader2023denoising} & $0.039\,(\pm\, 0.019)$ & $0.181\,(\pm\, 0.09)$ & $0.668\,(\pm\, 0.121)$ \\
\bottomrule
\end{tabular}
\caption{Comparison of generative models across three metrics: FID, F1 score, and coverage. \textsc{Form} outperforms baseline methods across all metrics. The arrows in the table represent performance metrics where a higher value indicates better performance.}
\label{table1}
\end{table}
\\
HA-GAN, while faster at inference due to its single-step generation, relies on patch-based learning to capture both local and global structure. It produced smooth samples in some cases, such as nocodazole (Figure~\ref{figure3}c), but struggled with fidelity and precision, likely due to the limited capacity to model fine morphological detail. MedicalDiffusion, though diffusion-based like \textsc{Form}, does not treat channels separately, leading to competitive diversity but reduced sample clarity, reflected in lower FID and qualitatively (Figure~\ref{figure3}c), possibly due to its neglect of spatial relationships between nucleus and cytoplasm structures that \textsc{Form} preserves. 
\\
Qualitatively, \textsc{Form}-generated samples exhibit higher visual realism than those of competing baselines (Figure~\ref{figure3}c). The synthesised volumes preserve realistic 3D structure and subcellular detail across axial, sagittal, and coronal views when rendered in Napari (Figure~\ref{figure3}b), reinforcing the biological plausibility of the generated cells.
\\
\subsection{FORM Predicts Perturbation-Specific Morphologies from Controls}\label{transition}

\begin{figure}[htbp]
    \centering
    \includegraphics[width=\linewidth]{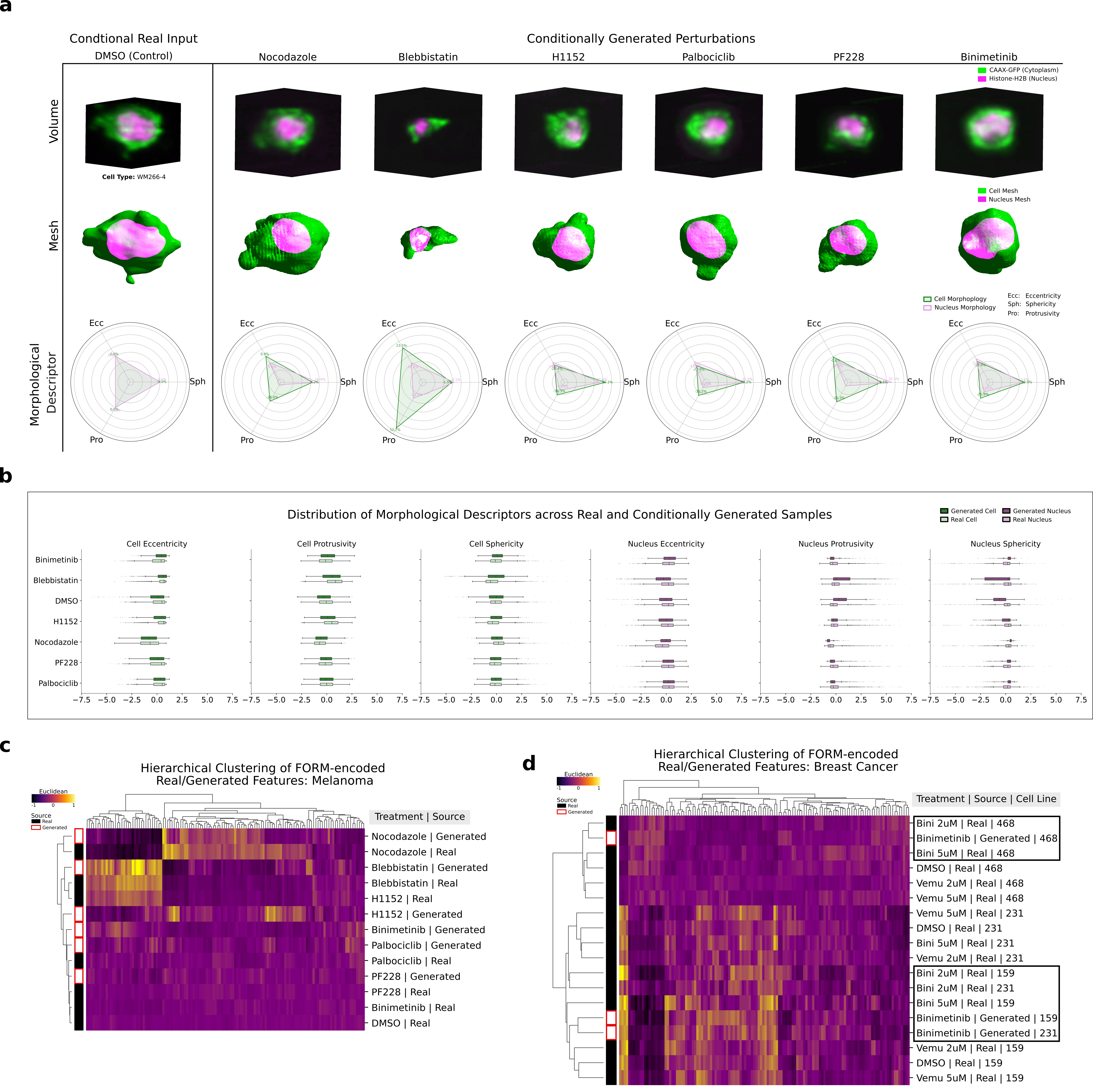}
    \caption{\textbf{Conditional generation with \textsc{Form}.}
    \textbf{a)} Conditional generation from an untreated (DMSO) input cell. The leftmost column shows the real control; subsequent columns show \textsc{Form}-generated post-treatment morphologies under different perturbation prompts. Top: 3D volumetric renderings. Middle: mesh reconstructions. Bottom: relative percentage change in key morphological descriptors versus the untreated control.
    \textbf{b)} Distributions of morphological descriptors across all generated samples ($N=1{,}000$ per perturbation) benchmarked against real counterparts ($N=1{,}000$).
    \textbf{c)} Hierarchical clustering of \textsc{Form} Encoder embeddings for real and generated samples across perturbations, showing that generated cells co-cluster with their corresponding real treatment groups.
    \textbf{d)} Cross-subtype generalisation: using the WM266-4 binimetinib model, DMSO controls from TNBC cell lines (231, 468, 159) were used as conditioning inputs to generate binimetinib-treated morphologies; a hierarchical clustermap of \textsc{Form} Encoder embeddings across all cell lines and perturbation conditions (real and generated) shows that generated samples co-cluster with their corresponding real groups.}
    \label{figure4}
\end{figure}

Building on \textsc{Form}'s capacity to stochastically generate diverse perturbation-specific morphologies, we next evaluate its performance in the \textit{conditional} setting. Here, the generative process is guided by an input cell and a specified treatment condition, enabling the synthesis of a plausible post-treatment morphology of the conditional input cell. This supports \textit{in silico} phenotype translation, where untreated cells are computationally mapped to their expected treatment-induced morphological states.
\\
\textsc{Form} acquires this capability without being trained on explicitly paired untreated–treated examples. Although recent literature has introduced methods that use stochastic differential equations to translate images between source and target distributions \citep{zheng2025diffusion,zhou2024denoising}, \textsc{Form} remains task-agnostic during training. Instead, by learning the intra-perturbation structure across a spectrum of individual treatments, the model captures distinct regions of the phenotypic landscape, enabling transitions between conditions to be inferred. This process can be interpreted as a morphological ``bridging'' mechanism guided by \textsc{Form}'s diffusion-based Perturbation Trajectory Module. Although diffusion models are not explicitly trained with directional supervision, the sequential nature of forward (noising) and reverse (denoising) steps imposes a structured progression through the latent space. 
\\
In our conditional setup, an untreated input cell $x_{\text{DMSO}}$ is first encoded and corrupted with Gaussian noise for $t$ steps, producing an intermediate representation $x_t$ that lies within a stochastic interpolation zone. The reverse diffusion process then denoises $x_t$ under the influence of a target treatment condition, progressively guiding the sample toward the morphological manifold associated with the perturbed phenotype. The resulting sample $x_{\text{Treated}}$ thus emerges as a plausible condition-aligned synthesis, bridging phenotypic distributions through a structured latent trajectory.
\\
To evaluate the conditionally generated samples, under the \textsc{MorphoEval} framework, we examine both the morphological changes induced by each perturbation and the extent to which these changes align with the morphological descriptors of real treatment-specific cell populations. Shape descriptors were extracted by first converting each generated 3D volume into a mesh object (detailed in Methods), from which we computed key morphological features for both the cytoplasm and nucleus channels. These features are visualised in Figure~\ref{figure4}a as relative changes from the control input. In doing so, \textsc{Form} not only generates treatment-specific morphologies, but also provides a quantifiable estimate of the expected shape change induced by a given perturbation. 
\\
Quantifying treatment-induced shape changes using classical morphological descriptors grounds our generative framework in biological interpretability. This enables direct validation of model predictions against well-characterised phenotypic outcomes. For instance, as shown in Figure~\ref{figure4}a, when a DMSO-treated cell is conditioned on blebbistatin, the model predicts a 52.7\% increase in cellular protrusivity alongside a marked decrease in sphericity, consistent with the expected spindly morphology induced by inhibition of myosin II \citep{Sero2015-ll}. In contrast, conditioning the same DMSO-treated cell on nocodazole results in a morphology with increased sphericity and a substantial reduction in protrusivity, reflecting the characteristic rounding associated with microtubule depolymerisation \citep{Sero2015-ll}. 
\\
To assess whether these morphological trends persisted across cell populations, we examined the distribution of generated descriptors at scale. Sampling from a population of DMSO-treated cells, we conditionally generated $1,000$ samples per perturbation and compared their morphological descriptors to an equally sized subset of real, treatment-specific cells. As shown in Figure~\ref{figure4}b, the distribution of descriptors from the generated samples closely matches that of the real cells, reinforcing the biological plausibility of the model's perturbation-effect predictions at scale. 
\\
To further validate these findings in a conditional setting, we assessed whether \textsc{Form}-generated samples capture perturbation-specific structure in feature space. Using the \textsc{Form} Encoder, we extracted embeddings from both real and generated volumes and constructed a hierarchical cluster map across perturbation conditions. We observed that generated samples consistently grouped alongside their corresponding real counterparts, indicating that \textsc{Form} preserves perturbation-specific morphological signatures rather than collapsing to generic cell-like structures.
\\

\subsection{FORM Generalises to Unseen Cancer Subtypes}\label{generalisation}

To evaluate whether \textsc{Form} can generalise beyond the training context, we applied the binimetinib-trained WM266-4 melanoma Perturbation Trajectory Module to a triple-negative breast cancer (TNBC) dataset comprising 468, 231, and 159 cell lines. Conditioned on DMSO-treated cells of each TNBC cell line, we generated corresponding binimetinib-treated morphologies. Feature embeddings extracted with the \textsc{Form} Encoder revealed that these generated samples clustered tightly with their respective TNBC cell line and perturbation groups, seen in Figure~\ref{figure4}d. This result demonstrates \textsc{Form}’s capacity to generalise across previously unseen cancer subtypes, producing perturbation-specific morphologies that remain consistent within distinct cellular contexts.
\\
\subsection{FORM Reveals Perturbation-induced Cellular Morphodynamics}\label{traversal}

\begin{figure}[htbp]
    \centering
    \includegraphics[width=\linewidth]{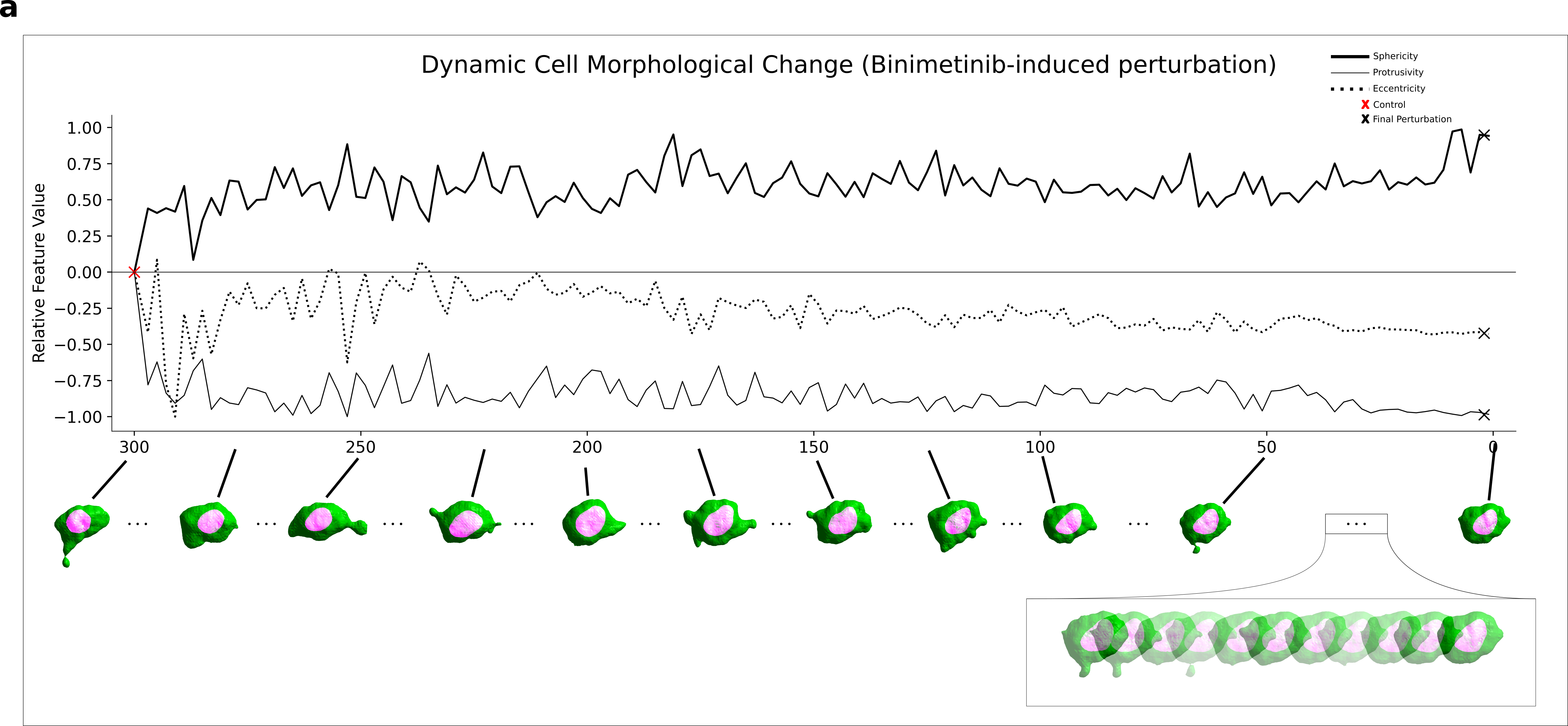}
    \caption{\textbf{Morphodynamic evolution of \textsc{Form}-generated cells.}  
    Morphological descriptor trajectories (eccentricity, sphericity, protrusivity, etc.) are shown as a function of denoising timestep ($t = 300 \rightarrow 0$), capturing how cellular shape evolves during the generative process. Below, representative 3D renderings of generated cells at selected timesteps illustrate the corresponding structural transitions, linking quantitative descriptor changes with visually interpretable morphology.}  
    \label{figure5}
\end{figure}

Quantifying how cellular morphology dynamically evolves under different perturbation conditions is central to morphological profiling. Traditional methods often rely on static comparisons or linearly interpolate between discrete treatment states \citep{Bou_PhenDiff_MICCAI2024,Palma2025,navidi2025morphodiff}, a strategy that risks oversimplifying the complex and often non-linear dynamics of morphological change. Although style- or content-based interpolation produces smooth transitions \citep{navidi2025morphodiff}, these approaches typically assume linear evolution in morphology, an assumption that may not accurately reflect biological reality. In living systems, morphological transformations are often sporadic, stochastic, and context-dependent \citep{Copperman2023}. To support this, we analysed the morphological dynamics of a live cell imaged at five-minute intervals over a ten-hour period. We show in the Supplementary Materials (Figure \ref{supp3}) that the resulting trajectory does not follow a smooth interpolation between phenotypic states. Instead, it reveals abrupt and heterogeneous shifts in shape, highlighting the limitations of linear interpolation-based assumptions in modelling cell state transitions.
\\
These observations motivate a more nuanced approach to modelling morphological dynamics. In this context, \textsc{Form} enables the in silico reconstruction of ``phenotypic traversals,'' offering a principled framework to observe and quantify treatment-induced morphodynamics through continuous diffusion trajectories. Via the morphological bridging mechanism described in Section \ref{transition}, an input cell undergoes a guided noising–denoising process that transforms it towards a perturbation-specific state. To better characterise how morphology evolves during this transformation, we track shape changes directly from the intermediate, progressively noised latent states. More specifically, given a conditional input sample $x_{DMSO}$, we first encode and progressively noise the input sample according to the control diffusion model. The resulting latent representation at the time step $T$ can be denoted as $x_{Treated, T}$ - a noisy sample that requires denoising over the $T$ steps to arrive at the final fully denoised phenotype $x_{Treated, 0}$. This endpoint represents the predicted post-treatment morphology. However, the trajectory from $x_{Treated, T}$ to $x_{Treated, 0}$ comprises a sequence of intermediate states at each time step $t \in\{1, T\}$. To analyse the dynamics of shape evolution, we treat each intermediate noisy latent $x_{Treated, t}$ as an initial condition and denoise it for exactly $t$ steps. This yields a series of denoised reconstructions that approximate the most probable morphological trajectory a cell might undergo under a given perturbation, effectively tracing the bounds of its expected shape evolution within the learnt phenotypic landscape (depicted in Figure \ref{figure5}). 
\\
To quantitatively evaluate the fidelity of \textsc{Form}-generated dynamics relative to true biological morphodynamics, we extracted Catch22 time-series features \citep{Lubba2019} from the real live cell's morphological evolution. We then computed the absolute differences between these features and those extracted from \textsc{Form}-simulated trajectories. For comparison, we also performed the same analysis on a linearly interpolated sequence generated between initial and final cell states. Our results (Supplementary Materials) show that \textsc{Form}-generated traversals more closely align with real dynamic morphological patterns, outperforming simple linear interpolation. In line with the conditional setup, each volumetric state along the trajectory is converted into a mesh, enabling the extraction of classical morphological descriptors at each step, thereby offering a principled approach to quantifying evolving 3D shape changes throughout the generative process. Taken together, these results demonstrate that \textsc{Form} provides a principled route for modelling continuous 3D morphodynamic transitions, moving beyond interpolation-based heuristics toward a generative framework that more faithfully reflects the stochastic and heterogeneous nature of live-cell shape evolution.
\\
\subsection{FORM Simulates Intracellular Signalling Activity}

\begin{figure}[H]
    \centering
    \includegraphics[width=\linewidth]{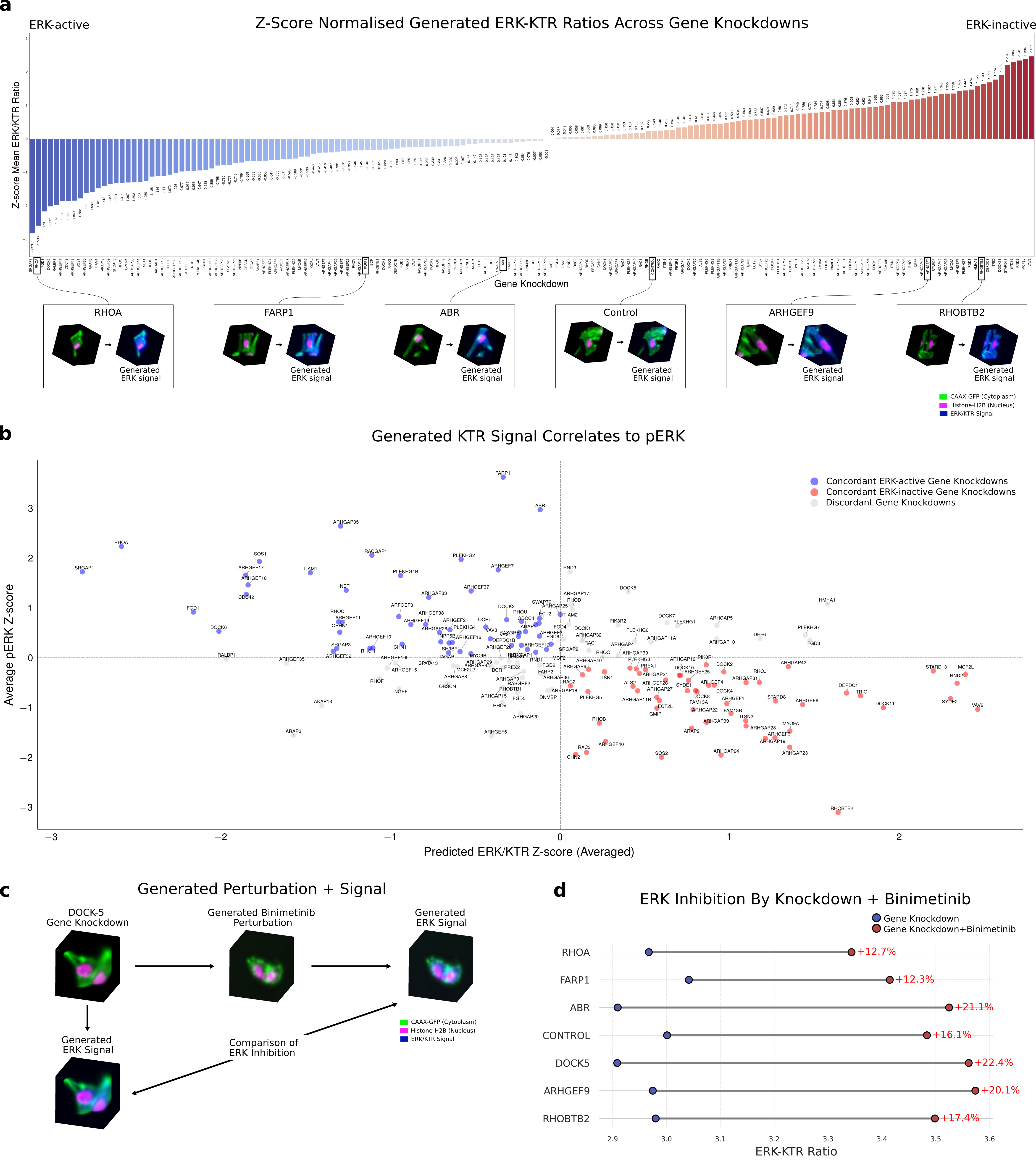}
    \caption{\textbf{FORM models perturbation-induced changes in ERK signalling activity.} 
    \textbf{a)} \textsc{Form} is used to conditionally generate KTR activity maps for each gene knockdown from the RNAi library. From these predictions, ERK-KTR ratios are calculated, averaged per knockdown, and Z-score normalised. Representative examples are shown, where cytoplasm and nucleus inputs are used to synthesise the corresponding ERK-KTR signal.
    \textbf{b)} Z-score normalised predicted ERK-KTR ratios are compared to experimentally measured pERK intensities across knockdowns. Each point represents a gene, illustrating the alignment between predicted signalling states and true biochemical measurements.
    \textbf{c)} Schematic of the simulation pipeline: untreated gene knockdowns are conditionally transformed to model the impact of binimetinib treatment, enabling the generation of predicted ERK activity maps.
    \textbf{d)} Lollipop plot showing the percentage increase in ERK-KTR ratio following simulated binimetinib treatment across a range of gene knockdowns, reflecting predicted ERK inhibition.
    }
    \label{figure6}
\end{figure}

Although \textsc{Form} models morphological transitions across perturbations, these transitions are often predictive of underlying intracellular signaling states \citep{DeVries2025,doi:10.1126/science.1140324,Yin2013,Cooper2015-jw,Sero2015-ll,Sailem2015,Sero2017,Way2022-bh,10.1098/rsob.130132}. Among these, the MAPK/ERK pathway plays a central role in the regulation of cell proliferation, differentiation, and drug response. ERK activity can be quantified using live-cell biosensors such as ERK-KTR \citep{Way2022-eo,Simpson306571,Kudo2018}, which translocate between the nucleus and cytoplasm depending on phosphorylation state, providing a dynamic readout of kinase signalling at the single-cell level.
\\
\textsc{Form} was retrained to generate the ERK-KTR signal directly from 3D cell and nuclear morphology, extending its generative capacity beyond structural synthesis to functional prediction. Although prior work has used morphological classifiers to infer perturbations associated with specific signalling pathways, such as MEK inhibition \citep{DeVries2025}, our approach adopts a generative framework. Rather than predicting pathway activity through classification scores, we synthesise the ERK-KTR signal as an image channel, conditioned on morphology, enabling spatially resolved prediction of intracellular kinase activity. ERK activity is quantified using the ERK ratio, where the mean intensity of ERK-KTR in the nucleus is divided by that in the surrounding nuclear ring, and a higher ratio indicates lower ERK signalling (see Supplementary materials for further details). 
\\
We evaluated this approach on the RNAi dataset by applying a \textsc{Form} model trained on the drug-treated WM266-4 cells. For each of the 167 gene knockdowns, \textsc{Form} generated ERK-KTR signals from cell and nuclear morphology. We computed ERK ratios per cell, averaged them per condition, and z-normalised the results to reveal knockdown-specific patterns of inferred ERK activity (Figure~\ref{figure6}a). To validate that the inferred ERK-KTR signals reflected true biochemical activity, we compared them to nuclear pERK levels measured via 2D immunofluorescence imaging from an independent RNAi screen using the same cell line and library (Figure~\ref{figure6}b). 
\\
Although \textsc{Form} predictions are based solely on morphology and pERK is measured biochemically, the model's output exhibited a moderate inverse Pearson correlation ($\rho$ = -0.50) with pERK (Figure~\ref{figure5}b). This inverse trend is consistent with the biological mechanism, where elevated cytoplasmic KTR typically corresponds to reduced nuclear pERK. Additionally, the predictions demonstrated a concordance index of 0.68, indicating strong agreement in the relative ranking of perturbation effects. A Kolmogorov–Smirnov (KS) test between the predicted and true z-score distributions did not show significant differences (KS statistic = 0.078, $p$ = 0.69), suggesting that the model also captures the global distribution of ERK activity.
\\
Finally, to explore whether \textsc{Form} can simulate perturbation-induced changes in kinase activity, we focused on a subset of gene knockdowns with known effects on ERK signalling. For each of these knockdowns, we employed \textsc{Form} to conditionally generate the corresponding binimetinib-treated morphology, effectively simulating how each genetic background responds to MEK inhibition (Figure \ref{figure6}c). By generating ERK-KTR activity readouts for both the unperturbed and simulated perturbation states, we could investigate whether predicted kinase activity patterns aligned with biological expectations. In particular, whether ERK-inactive knockdowns showed a larger degree of suppressed signalling after binimetinib treatment, and whether ERK-active states showcased marginal ERK-inhibition. Our findings (Figure~\ref{figure6}d) align with this expectation. RHOBTB2 and ARHGEF9 exhibit an approximate 7\% larger inhibition in ERK than RHOA and FARP1. Notably, \textsc{Form} predicts that DOCK5 knockdown in combination with binimetinib treatment yields the strongest ERK inhibition across the tested conditions (Figure~\ref{figure6}d). This aligns with previous experimental work showing that LM2 cells, which are typically resistant to MEK inhibition, become highly sensitive when DOCK5 is depleted \citep{pascual2021multiplexed}. \textsc{Form} successfully recapitulates this known synergistic effect, suggesting that it captures not only morphological responses but also genotype-specific treatment vulnerabilities.

\section{Discussion}

This work introduces \textsc{Form}, the first generative framework capable of simulating biologically realistic 3D single-cell morphologies across drug perturbations. Whereas prior approaches have largely been restricted to discriminative analyses or 2D image synthesis, \textsc{Form} establishes in silico perturbation modelling in three dimensions --- capturing both structural and functional cellular responses. 
\\
We show that \textsc{Form} predicts post-treatment morphologies directly from untreated controls, preserving perturbation-specific shape descriptors and supporting virtual phenotype translation. These predictions remain coherent across distinct cancer subtypes and cell lines, demonstrating that the framework generalises beyond its training context. Together, these results establish \textsc{Form} as a generative analogue to perturbation assays, with potential for testing drug responses in otherwise inaccessible settings.
\\
Beyond static synthesis, \textsc{Form} proposes a principled approach for modelling morphodynamic change. By tracing descriptor trajectories across denoised interpolation states, the framework provides a continuous view of cell-shape adaptation under perturbation. Although our implementation is an initial proof of concept, this paradigm, using generative trajectories to approximate live-cell shape evolution, offers a foundation for future efforts to capture aberrant dynamics directly from static imaging datasets.
\\
Finally, \textsc{Form} demonstrates that morphological simulation can be extended to intracellular signalling. By conditioning on structure to generate ERK-KTR activity, the framework unifies morphological and biochemical phenotypes within a single model. It recapitulates orthogonal pERK measurements and recovers context-specific patterns of kinase inhibition under genetic–drug combinations, supporting in silico exploration of genotype–treatment interactions.
\\
While the present study focuses on selected perturbations and cell types, broadening the framework across cellular systems and signalling pathways will be essential to establish its generality. Nonetheless, these results highlight how 3D generative modelling can bridge morphology, dynamics, and signalling, laying the foundation for virtual cell models to drive mechanistic insight, hypothesis generation, and drug discovery. 

\newpage

\section{Online Methods}

\subsection{Model}

\textbf{Overview of FORM}

The \textsc{Form} framework comprises a two-stage generative pipeline for synthesising drug-perturbed 3D cellular morphologies. The first stage involves learning a discrete latent representation of 3D cellular structures through vector quantisation, while the second stage facilitates the generation of new morphologies using a multichannel denoising diffusion model. A separate \textsc{Form} model is trained for each drug perturbation, allowing independent representation learning for different treatment conditions. 
\\
The pipeline is designed to capture cellular heterogeneity at a subcellular level by independently encoding the cytoplasm and nucleus. To achieve this, \textsc{Form} employs a library of vector quantised codebooks, where distinct learnable dictionaries store morphological features for each subcellular compartment. These learned representations serve as compressed latent descriptors, which are then passed through a latent diffusion model to generate high-resolution 3D cellular structures that simulate perturbation effects.
\\
\textbf{Vector Quantisation and Codebook Learning}

To establish a structured and discrete representation of cellular morphology, \textsc{Form} employs a Vector-Quantised Generative Adversarial Network (VQGAN) for each perturbation. This stage maps volumetric cellular data into a compressed latent space while enforcing a discretisation step to encourage structured feature learning. 
\\
Each 3D volume, denoted as $x \in \mathbb{R}^{C \times H \times W \times D}$, consists of two channels: the cytoplasmic membrane and the nuclear compartment, which are processed separately. The volume is decomposed into its respective channels, represented as $x_{cell} \in \mathbb{R}^{1 \times H \times W \times D}$ and $x_{nuc} \in \mathbb{R}^{1 \times H \times W \times D}$, which are independently encoded via a 3D convolutional encoder, $E$, yielding latent representations:
\begin{equation}
\hat{z}_{cell} = E(x_{cell}), \quad \hat{z}_{nuc} = E(x_{nuc})
\end{equation}
where $\hat{z}_{cell} \in \mathbb{R}^{1 \times h \times w \times n_z}$ and $\hat{z}_{nuc} \in \mathbb{R}^{1 \times h \times w \times n_v}$ represent the encoded feature maps of the cell and nucleus, respectively, with $h < H$, $w < W$, and $n_z, n_v < D$ indicating spatial compression.

A vector quantisation step follows, where each latent vector is mapped to the closest entry in a discrete, learnable codebook:
\begin{equation} 
z_{q_{cell}} = \mathbf{q}(\hat{z}_{cell}) = \arg\min_{z_k \in Z_{cell}} ||\hat{z}_{ij} - z_k|| 
\end{equation}
\begin{equation} 
z_{q_{nuc}} = \mathbf{q}(\hat{z}_{nuc}) = \arg\min_{z_p \in Z_{nuc}} ||\hat{z}_{mn} - z_p|| 
\end{equation}
where $Z_{cell} = \{z_k\}_{k=1}^{K} \in \mathbb{R}^{1 \times n_z}$ and $Z_{nuc} = \{z_p\}_{p=1}^{P} \in \mathbb{R}^{1 \times n_v}$ are the learnable codebooks containing discrete embeddings for the cytoplasm and nucleus, respectively. Each entry in these codebooks represents a prototypical morphological feature, enabling cellular structures to be represented as spatial arrangements of a finite set of learned features.

Once quantised, the latent representations are decoded via a 3D convolutional decoder $G$, reconstructing the full 3D volume:
\begin{equation}
\hat{x} = G(z_{q_{cell}}, z_{q_{nuc}}) = G(\mathbf{q}(E(x_{cell})), \mathbf{q}(E(x_{nuc})))
\end{equation}

To ensure stable and high-quality learning, the VQGAN is optimised using three key loss functions:
\\
\textbf{Reconstruction Loss ($L_{rec}$)} ensures that the generated output $\hat{x}$ closely matches the input volume $x$ by minimising the pixel-wise reconstruction error:
\begin{equation} 
L_{rec} = \frac{1}{2} \left[ ||x_{cell} - \hat{x}_{cell}||^2 + ||x_{nuc} - \hat{x}_{nuc}||^2 \right]
\end{equation}
\\
\textbf{Commitment Loss ($L_{comm}$)} encourages the encoded representation to stay close to the quantised codebook entry to ensure stability in latent space representation:
\begin{equation} 
L_{comm} = \frac{1}{2} \left[ ||\text{sg}[z_{q_{cell}}] - E(x_{cell}) ||_2^2 +  ||\text{sg}[z_{q_{nuc}}] - E(x_{nuc}) ||_2^2\right] 
\end{equation}
where $\text{sg}$ is the stop-gradient operation, which prevents the encoder from receiving gradients from the quantisation operation, ensuring that the quantised embeddings are learned independently.
\\
\textbf{Adversarial Loss ($L_{disc}$)} improves the perceptual quality of the generated samples by incorporating a discriminator $D$ trained to distinguish between real and generated cellular structures:
\begin{equation} 
L_{disc} = \frac{1}{2} \left[ \mathbb{E}_{x}(\text{ReLU}(1-D(x)) + \mathbb{E}_{\hat{x}}(\text{ReLU}(1-D(\hat{x})))) \right] 
\end{equation}
where $D(x)$ and $D(\hat{x})$ represent the discriminator predictions for real and generated samples, respectively. This loss encourages the generator to synthesise realistic cellular structures by learning a structured mapping of perturbation-induced morphologies.

By jointly optimising these loss functions, the VQGAN effectively learns to encode, quantise, and reconstruct 3D cellular structures, forming a robust foundation for generative modelling in \textsc{Form}.
\\
\textbf{Multichannel Denoising Diffusion Modelling}

Each \textsc{Form} model is trained independently for a given perturbation setting, with the diffusion model learning perturbation-conditioned generative processes within the unquantised latent space defined by the VQGAN. The diffusion model enables controlled sampling within this latent space, modelling how morphological transitions occur across perturbation conditions.

The forward diffusion process applies a controlled stochastic transformation to latent representations, gradually adding Gaussian noise:
\begin{equation} 
q(\hat{z}_t | \hat{z}_{t-1}) = \mathcal{N}(\hat{z}_t; \sqrt{1-\beta_t} \hat{z}_{t-1}, \beta_t \mathbf{I})
\end{equation}
where $\beta_t$ defines a variance schedule that progressively increases over diffusion steps $T$. This process ensures that samples eventually converge to a Gaussian prior, from which novel perturbation-conditioned latent representations can be generated.

The reverse diffusion process learns to denoise and generate structured latent representations, thereby enabling sampling from perturbation distributions:
\begin{equation}
p_\theta(\hat{z}_{t-1} | \hat{z}_t) = \mathcal{N}(\hat{z}_{t-1} ; \mu_\theta(\hat{z}_t, t), \sigma_\theta^2(\hat{z}_t, t))
\end{equation}
where $\mu_\theta$ and $\sigma_\theta^2$ are neural network parameterised functions predicting the denoised latent at each step.

To model this, we employ a dual-channel UNet, an adapted 3D UNet architecture specifically designed to handle multichannel diffusion processes. The dual-channel UNet simultaneously processes latent representations of both cytoplasm and nucleus, enforcing spatial consistency between subcellular components. The architecture incorporates spatial- and depth-wise attention mechanisms, ensuring that features across both channels interact meaningfully while preserving fine-grained morphological details. The final sampled latent representations are then decoded via the pre-trained VQGAN decoder, producing high-fidelity 3D cellular structures that reflect treatment-induced morphological variation.

\subsection{Datasets}

This study used four internally generated \citep{DeVries2025} datasets : (1) a small-molecule screen of WM266-4 melanoma cells embedded in collagen and imaged using stage-scanning oblique plane microscopy (ssOPM), (2) a triple-negative breast cancer (TNBC) dataset imaged using a ssOPM, (3) an RNAi screen of the same cells imaged on ssOPM, and (4) a pERK RNAi screen of WM266-4 cells imaged in 2D using the Opera QEHS platform.
\\
WM266-4 cells were genetically modified to express CAAX-EGFP, ERK-KTR-Ruby (Addgene \#90231), and H2B-iRFP670 (Addgene \#90237). Cells were embedded in 2mg/mL collagen hydrogels and seeded at 40,000 cells per well in 96-well plates. After 24 hours, cells were treated with various compounds (binimetinib, blebbistatin, nocodazole, CK666, H1152, PF228, MK1775) for 6 hours and fixed with 4\% PFA. Final concentrations were adjusted to account for hydrogel volume. 3D imaging was performed using ssOPM.
\\
TNBC 159, 468, and 231 cells were embedded in 2 mg/mL collagen hydrogels and seeded in 96-well plates at the same density as WM266-4 cells. After 24 h, cells were treated with binimetinib or vemurafenib at multiple concentrations, fixed with 4\% PFA, and imaged in 3D using ssOPM.
\\
For the pERK RNAi screen, WM266-4 cells were reverse-transfected in 384-well plates with 168 siRNA conditions from a custom RhoGEF/RhoGAP \citep{Bousgouni2022-pn} library using ON-TARGETplus SmartPools (Dharmacon). After 48 hours, cells were fixed and stained for pERK, actin, and DNA, and imaged in a single 2D plane using the Opera QEHS system with a 20x objective.

\subsection{Data Processing}

\textbf{Volume Preparation for Modelling.} 

Each 3D single-cell volume, comprising stacked cytoplasm and nucleus channels, is rescaled to a fixed size of $64 \times 64 \times 64$ using isotropic resizing followed by zero-padding as needed. The resulting volumes are normalised to the range $[-1, 1]$ to stabilise training and improve convergence in diffusion-based generative modelling.
\\
\textbf{Mesh Construction at Inference.} 

To enable quantitative assessment of generated cell shape, we transformed the output volumes into 3D surface meshes. While voxel-based representations are suitable for training, downstream morphological descriptors --- such as surface area, sphericity, and protrusivity --- are best computed on smooth, continuous surfaces. Mesh-based representations not only support this analysis, but also provide clearer visualisations of structural detail.
For each generated sample, the cytoplasm and nucleus channels were separately thresholded using Otsu's method to extract a binary boundary. The marching cubes algorithm \citep{10.1145/37401.37422,Lewiner01012003} from scikit-learn \citep{10.5555/1953048.2078195,buitinck2013apidesignmachinelearning} was then applied to extract surface geometry from each channel, producing vertices and faces corresponding to the predicted morphological boundaries.

\subsection{Baselines}

\textbf{HA-GAN. } 

HA-GAN \citep{hagan} is a GAN-based architecture designed to synthesise high-resolution 3D images while mitigating the memory constraints of volumetric data. In their original experiments, the authors evaluated HA-GAN on 3D brain (GSP \citep{Holmes2015}) and lung (COPDGene \citep{Regan2010-ul}) MRI and CT datasets. During training, HA-GAN generates a low-resolution full image and a randomly selected high-resolution sub-volume. This hierarchical structure preserves morphological consistency across the volume while enabling learning of fine-grained features. During inference, the model synthesises entire high-resolution volumes in a single pass. We adapted HA-GAN for our application by adjusting its input resolution to match the ${64}^3$ voxel format and training it on the same treatment-specific subsets used in \textsc{Form}.

\textbf{MedicalDiffusion.} 

MedicalDiffusion \citep{khader2023denoising} is a diffusion-based model developed for synthesising medical images. The original MedicalDiffusion model was trained on publicly available 3D datasets spanning four anatomical regions: brain MRI (ADNI \citep{Petersen2009-bh}), chest CT (LIDC \citep{Armato2015}), breast MRI (DUKE \citep{Saha2021}), and knee MRI (MRNet \citep{10.1371/journal.pmed.1002699}). It learns to map Gaussian noise to high-resolution 3D images by inverting a noising process through a UNet-based architecture. Unlike \textsc{Form}, which encodes cytoplasm and nucleus channels separately, MedicalDiffusion models both jointly as a single input tensor. We trained this model using identical noise schedules and data splits for comparability. 

\subsection{Quantitative Metrics}

\textbf{Fréchet Inception Distance (FID).}

The Fréchet Inception Distance (FID) \citep{7780677} quantifies the distance between real and generated data distributions in a learned feature space. Conventionally, FID is computed by extracting features from the penultimate layer of an InceptionV3 \citep{7780677} network trained on ImageNet \citep{5206848}, providing a perceptual embedding of each image. The statistics (mean and covariance) of these embeddings are then compared under the assumption that both real and generated features follow multivariate Gaussian distributions. Let $\mu_r, \mu_g$ and $C_r, C_g$ denote the means and covariances of the real and generated distributions, respectively. The FID is then computed as:
\begin{align}
FID = ||\mu_r - \mu_g||^2+Tr\bigg( C_r + C_g-2(C_r C_g)^{1/2}\bigg).
\end{align}

While FID is widely used in natural image synthesis, it is suboptimal for evaluating biological volumes, which differ markedly in structure and content from ImageNet images. To this end, we adapted the FID metric for our 3D volumetric data by extracting features using the pretrained \textsc{Form} encoder, trained directly on 3D cellular morphologies. This domain-specific encoder produces meaningful embeddings aligned with biological variation, enabling a more faithful comparison of generated and real samples. We compute FID using these embeddings, measuring both fidelity and distributional similarity in the morphological latent space.
\\
\textbf{Coverage \& F1 Score.}

In addition to FID, and inspired by the evaluation contributions of Palma et al. \citep{Palma2025}, we use geometric distribution-based metrics to evaluate the fidelity and diversity of generated 3D cellular morphologies. 

Let $\mathcal{R}= \{ r_1, r_2, ..., r_n\}$ be the set of real cell embeddings, and $\mathcal{G} = \{ g_1, g_2, ..., g_m\}$ be the set of generated cell embeddings, where each $r_i, g_j \in \mathbb{R}^d$ is a feature vector in a $d$-dimensional embedding space.

For each embedding, we compute its Euclidean distance to its $k$-nearest neighbours within its own set to define a local support radius. 

\textit{Precision} quantifies realism, defined as the fraction of generated samples $g_j \in \mathcal{G}$ that lie within the support radius of at least one real sample. \textit{Recall} quantifies diversity, defined as the fraction of real samples $r_i \in \mathcal{R}$ that lie within the support of at least one generated sample.

We report the harmonic mean of these two quantities as the $F_1$ score:

\begin{align}
F_1 = \frac{2 \cdot \text{Precision} \cdot \text{Recall}}{\text{Precision} + \text{Recall}}.
\end{align}

\textit{Coverage} provides a complementary measure of diversity. For each real sample $r_i$, we define a sphere centered at $r_i$ with radius equal to its distance to its $k$-th nearest real neighbour. Coverage is defined as the fraction of real samples whose sphere contains at least one generated sample. A high coverage score indicates that the generated distribution spans the full morphology space of the real data.
\\
\textbf{Concordance Index.}

To assess rank agreement between predicted and true ERK activity across perturbations, we compute the concordance index (CI): the probability that, for a randomly selected pair of conditions, the ordering of predicted values matches the ordering of ground truth. Since higher ERK/KTR ratios indicate lower signalling, we negate the predicted values before computing CI. Formally, for a set of $n$ paired observations, ${(x_i, \hat{x_i})}$ where $x_i$ are the ground truth scores and $\hat{x_i}$ are the predicted scores, CI is defined:

\begin{align}
CI = \frac{1}{N}\sum_{i<j} I\bigg[(x_i > x_j) \cap (\hat{x_i} > \hat{x_j})\bigg], 
\end{align}

where $N$ is the number of comparable pairs, and $I$ is the indicator function. A CI of $1.0$ indicates perfect ordering, while $0.5$ implies pure random ordering.

\subsection{ERK-KTR Ratio Measurements.}

Nuclear ERK-KTR intensity was quantified as the mean signal within the nucleus mask, calculated as: 

\begin{align}
\text{ERK Ratio} = \frac{\text{Mean Nuclear ERK Intensity}}{\text{Mean Ring Region ERK Intensity}}, 
\end{align}
where a ring region is obtained by expanding the nuclear mask via binary dilation of $7$ iterations.

\backmatter

\newpage

\bmhead{Supplementary information}

Supplementary figures have been attached. 

\bmhead{Acknowledgements}

This is a summary of independent research supported by the National Institute for Health Research (NIHR) Biomedical Research Centre at The Royal Marsden NHS Foundation Trust and The Institute of Cancer Research. The views expressed are those of the author(s) and not necessarily those of the NHS, the NIHR or the Department of Health and Social Care. The Authors would also like to thank StratMedPaediatrics2 funded by Cancer Research UK, CRCEMA-Jul23/100001 

\bmhead{Author Contributions}

Conceptualisation, R.N., and C.B.; methodology, R.N., and M.D.V.; investigation, R.N.; writing - original draft, R.N., and C.B.; writing - review and editing, R.N., M.D.V, O.F.,  V.B., M.A.G., M.P.M. and C.B.; funding acquisition, C.B.; resources, C.B.; supervision, C.B.

\bmhead{Declaration of Generative AI and AI-assisted Technologies}

ChatGPT was used to reword sentences. Authors reviewed and edited the content after use of this tool, and take full responsibility for the content of the publication.


\bibliography{sn-bibliography}

\begin{thebibliography}{10}
\expandafter\ifx\csname url\endcsname\relax
  \def\url#1{\burl{#1}}\fi
\expandafter\ifx\csname urlprefix\endcsname\relax\def\urlprefix{URL }\fi
\providecommand{\bibinfo}[2]{#2}
\providecommand{\eprint}[2][]{\url{#2}}
\providecommand{\doi}[1]{\url{https://doi.org/#1}}
\bibcommenthead

\bibitem{doi:10.1126/science.1140324}
\bibinfo{author}{Bakal, C.}, \bibinfo{author}{Aach, J.},
  \bibinfo{author}{Church, G.} \& \bibinfo{author}{Perrimon, N.}
\newblock \bibinfo{title}{Quantitative morphological signatures define local
  signaling networks regulating cell morphology}.
\newblock \emph{\bibinfo{journal}{Science}} \textbf{\bibinfo{volume}{316}},
  \bibinfo{pages}{1753--1756} (\bibinfo{year}{2007}).
\newblock
  \urlprefix\url{https://www.science.org/doi/abs/10.1126/science.1140324}.

\bibitem{doi:10.1126/science.aba2894}
\bibinfo{author}{Lomakin, A.~J.} \emph{et~al.}
\newblock \bibinfo{title}{The nucleus acts as a ruler tailoring cell responses
  to spatial constraints}.
\newblock \emph{\bibinfo{journal}{Science}} \textbf{\bibinfo{volume}{370}},
  \bibinfo{pages}{eaba2894} (\bibinfo{year}{2020}).
\newblock
  \urlprefix\url{https://www.science.org/doi/abs/10.1126/science.aba2894}.

\bibitem{Verde1995-nh}
\bibinfo{author}{Verde, F.}, \bibinfo{author}{Mata, J.} \&
  \bibinfo{author}{Nurse, P.}
\newblock \bibinfo{title}{Fission yeast cell morphogenesis: identification of
  new genes and analysis of their role during the cell cycle}.
\newblock \emph{\bibinfo{journal}{J Cell Biol}} \textbf{\bibinfo{volume}{131}},
  \bibinfo{pages}{1529--1538} (\bibinfo{year}{1995}).

\bibitem{noutahi2025virtualcellspredictexplain}
\bibinfo{author}{Noutahi, E.} \emph{et~al.}
\newblock \bibinfo{title}{Virtual cells: Predict, explain, discover}
  (\bibinfo{year}{2025}).
\newblock \urlprefix\url{https://arxiv.org/abs/2505.14613}.
\newblock
  \bibinfo{eprint}{{\href{https://arxiv.org/abs/2505.14613}{{arXiv:2505.14613}}}}.

\bibitem{Palma2025}
\bibinfo{author}{Palma, A.}, \bibinfo{author}{Theis, F.~J.} \&
  \bibinfo{author}{Lotfollahi, M.}
\newblock \bibinfo{title}{Predicting cell morphological responses to
  perturbations using generative modeling}.
\newblock \emph{\bibinfo{journal}{Nature Communications}}
  \textbf{\bibinfo{volume}{16}}, \bibinfo{pages}{505} (\bibinfo{year}{2025}).
\newblock \urlprefix\url{https://doi.org/10.1038/s41467-024-55707-8}.

\bibitem{Bou_PhenDiff_MICCAI2024}
\bibinfo{author}{Bourou, A.} \emph{et~al.}
\newblock \emph{\bibinfo{title}{{ PhenDiff: Revealing Subtle Phenotypes with
  Diffusion Models in Real Images }}}, Vol. \bibinfo{volume}{LNCS 15003}
  (\bibinfo{publisher}{Springer Nature Switzerland}, \bibinfo{year}{2024}).

\bibitem{Lamiable2023}
\bibinfo{author}{Lamiable, A.} \emph{et~al.}
\newblock \bibinfo{title}{Revealing invisible cell phenotypes with conditional
  generative modeling}.
\newblock \emph{\bibinfo{journal}{Nature Communications}}
  \textbf{\bibinfo{volume}{14}}, \bibinfo{pages}{6386} (\bibinfo{year}{2023}).
\newblock \urlprefix\url{https://doi.org/10.1038/s41467-023-42124-6}.

\bibitem{Cui2024}
\bibinfo{author}{Cui, H.} \emph{et~al.}
\newblock \bibinfo{title}{scgpt: toward building a foundation model for
  single-cell multi-omics using generative ai}.
\newblock \emph{\bibinfo{journal}{Nature Methods}}
  \textbf{\bibinfo{volume}{21}}, \bibinfo{pages}{1470--1480}
  (\bibinfo{year}{2024}).
\newblock \urlprefix\url{https://doi.org/10.1038/s41592-024-02201-0}.

\bibitem{Adduri2025}
\bibinfo{author}{Adduri, A.~K.} \emph{et~al.}
\newblock \bibinfo{title}{Predicting cellular responses to perturbation across
  diverse contexts with state}.
\newblock \emph{\bibinfo{journal}{bioRxiv}}  (\bibinfo{year}{2025}).
\newblock
  \urlprefix\url{https://www.biorxiv.org/content/early/2025/07/10/2025.06.26.661135}.

\bibitem{zhang2025cellfluxsimulatingcellularmorphology}
\bibinfo{author}{Zhang, Y.} \emph{et~al.}
\newblock \bibinfo{title}{Cellflux: Simulating cellular morphology changes via
  flow matching} (\bibinfo{year}{2025}).
\newblock \urlprefix\url{https://arxiv.org/abs/2502.09775}.
\newblock
  \bibinfo{eprint}{{\href{https://arxiv.org/abs/2502.09775}{{arXiv:2502.09775}}}}.

\bibitem{Mertz:19}
\bibinfo{author}{Mertz, J.}
\newblock \bibinfo{title}{Strategies for volumetric imaging with a fluorescence
  microscope}.
\newblock \emph{\bibinfo{journal}{Optica}} \textbf{\bibinfo{volume}{6}},
  \bibinfo{pages}{1261--1268} (\bibinfo{year}{2019}).
\newblock
  \urlprefix\url{https://opg.optica.org/optica/abstract.cfm?URI=optica-6-10-1261}.

\bibitem{Fischer2011-ql}
\bibinfo{author}{Fischer, R.~S.}, \bibinfo{author}{Wu, Y.},
  \bibinfo{author}{Kanchanawong, P.}, \bibinfo{author}{Shroff, H.} \&
  \bibinfo{author}{Waterman, C.~M.}
\newblock \bibinfo{title}{Microscopy in 3d: a biologist's toolbox}.
\newblock \emph{\bibinfo{journal}{Trends Cell Biol}}
  \textbf{\bibinfo{volume}{21}}, \bibinfo{pages}{682--691}
  (\bibinfo{year}{2011}).

\bibitem{DeVries2025}
\bibinfo{author}{De~Vries, M.} \emph{et~al.}
\newblock \bibinfo{title}{Geometric deep learning and multiple-instance
  learning for 3d cell-shape profiling}.
\newblock \emph{\bibinfo{journal}{Cell Systems}} \textbf{\bibinfo{volume}{16}}
  (\bibinfo{year}{2025}).
\newblock \urlprefix\url{https://doi.org/10.1016/j.cels.2025.101229}.

\bibitem{Roohani2024}
\bibinfo{author}{Roohani, Y.}, \bibinfo{author}{Huang, K.} \&
  \bibinfo{author}{Leskovec, J.}
\newblock \bibinfo{title}{Predicting transcriptional outcomes of novel
  multigene perturbations with gears}.
\newblock \emph{\bibinfo{journal}{Nature Biotechnology}}
  \textbf{\bibinfo{volume}{42}}, \bibinfo{pages}{927--935}
  (\bibinfo{year}{2024}).
\newblock \urlprefix\url{https://doi.org/10.1038/s41587-023-01905-6}.

\bibitem{Bunne2024}
\bibinfo{author}{Bunne, C.}, \bibinfo{author}{Schiebinger, G.},
  \bibinfo{author}{Krause, A.}, \bibinfo{author}{Regev, A.} \&
  \bibinfo{author}{Cuturi, M.}
\newblock \bibinfo{title}{Optimal transport for single-cell and spatial omics}.
\newblock \emph{\bibinfo{journal}{Nature Reviews Methods Primers}}
  \textbf{\bibinfo{volume}{4}}, \bibinfo{pages}{58} (\bibinfo{year}{2024}).
\newblock \urlprefix\url{https://doi.org/10.1038/s43586-024-00334-2}.

\bibitem{wu2025perturbenchbenchmarkingmachinelearning}
\bibinfo{author}{Wu, Y.} \emph{et~al.}
\newblock \bibinfo{title}{Perturbench: Benchmarking machine learning models for
  cellular perturbation analysis} (\bibinfo{year}{2025}).
\newblock \urlprefix\url{https://arxiv.org/abs/2408.10609}.
\newblock
  \bibinfo{eprint}{{\href{https://arxiv.org/abs/2408.10609}{{arXiv:2408.10609}}}}.

\bibitem{esser2021tamingtransformershighresolutionimage}
\bibinfo{author}{Esser, P.}, \bibinfo{author}{Rombach, R.} \&
  \bibinfo{author}{Ommer, B.}
\newblock \bibinfo{title}{Taming transformers for high-resolution image
  synthesis} (\bibinfo{year}{2021}).
\newblock \urlprefix\url{https://arxiv.org/abs/2012.09841}.
\newblock
  \bibinfo{eprint}{{\href{https://arxiv.org/abs/2012.09841}{{arXiv:2012.09841}}}}.

\bibitem{ddpm}
\bibinfo{author}{Ho, J.}, \bibinfo{author}{Jain, A.} \&
  \bibinfo{author}{Abbeel, P.}
\newblock \bibinfo{title}{Denoising diffusion probabilistic models}
  (\bibinfo{year}{2020}).
\newblock \urlprefix\url{https://arxiv.org/abs/2006.11239}.
\newblock
  \bibinfo{eprint}{{\href{https://arxiv.org/abs/2006.11239}{{arXiv:2006.11239}}}}.

\bibitem{khader2023denoising}
\bibinfo{author}{Khader, F.} \emph{et~al.}
\newblock \bibinfo{title}{Denoising diffusion probabilistic models for 3d
  medical image generation}.
\newblock \emph{\bibinfo{journal}{Scientific Reports}}
  \textbf{\bibinfo{volume}{13}}, \bibinfo{pages}{7303} (\bibinfo{year}{2023}).

\bibitem{Wang2025}
\bibinfo{author}{Wang, Z.} \emph{et~al.}
\newblock \bibinfo{title}{Image-based generation for molecule design with
  sketchmol}.
\newblock \emph{\bibinfo{journal}{Nature Machine Intelligence}}
  \textbf{\bibinfo{volume}{7}}, \bibinfo{pages}{244--255}
  (\bibinfo{year}{2025}).
\newblock \urlprefix\url{https://doi.org/10.1038/s42256-025-00982-3}.

\bibitem{unet}
\bibinfo{author}{Ronneberger, O.}, \bibinfo{author}{Fischer, P.} \&
  \bibinfo{author}{Brox, T.}
\newblock \bibinfo{editor}{Navab, N.}, \bibinfo{editor}{Hornegger, J.},
  \bibinfo{editor}{Wells, W.~M.} \& \bibinfo{editor}{Frangi, A.~F.} (eds)
  \emph{\bibinfo{title}{U-net: Convolutional networks for biomedical image
  segmentation}}.
\newblock (eds \bibinfo{editor}{Navab, N.}, \bibinfo{editor}{Hornegger, J.},
  \bibinfo{editor}{Wells, W.~M.} \& \bibinfo{editor}{Frangi, A.~F.})
  \emph{\bibinfo{booktitle}{Medical Image Computing and Computer-Assisted
  Intervention -- MICCAI 2015}}, \bibinfo{pages}{234--241}
  (\bibinfo{publisher}{Springer International Publishing},
  \bibinfo{address}{Cham}, \bibinfo{year}{2015}).

\bibitem{3dunet}
\bibinfo{author}{Özgün Çiçek}, \bibinfo{author}{Abdulkadir, A.},
  \bibinfo{author}{Lienkamp, S.~S.}, \bibinfo{author}{Brox, T.} \&
  \bibinfo{author}{Ronneberger, O.}
\newblock \bibinfo{title}{3d u-net: Learning dense volumetric segmentation from
  sparse annotation} (\bibinfo{year}{2016}).
\newblock \urlprefix\url{https://arxiv.org/abs/1606.06650}.
\newblock
  \bibinfo{eprint}{{\href{https://arxiv.org/abs/1606.06650}{{arXiv:1606.06650}}}}.

\bibitem{ddim}
\bibinfo{author}{Song, J.}, \bibinfo{author}{Meng, C.} \&
  \bibinfo{author}{Ermon, S.}
\newblock \bibinfo{title}{Denoising diffusion implicit models}
  (\bibinfo{year}{2022}).
\newblock \urlprefix\url{https://arxiv.org/abs/2010.02502}.
\newblock
  \bibinfo{eprint}{{\href{https://arxiv.org/abs/2010.02502}{{arXiv:2010.02502}}}}.

\bibitem{improvedddim}
\bibinfo{author}{Nichol, A.} \& \bibinfo{author}{Dhariwal, P.}
\newblock \bibinfo{title}{Improved denoising diffusion probabilistic models}
  (\bibinfo{year}{2021}).
\newblock \urlprefix\url{https://arxiv.org/abs/2102.09672}.
\newblock
  \bibinfo{eprint}{{\href{https://arxiv.org/abs/2102.09672}{{arXiv:2102.09672}}}}.

\bibitem{Bousgouni2022-pn}
\bibinfo{author}{Bousgouni, V.} \emph{et~al.}
\newblock \bibinfo{title}{{ARHGEF9} regulates melanoma morphogenesis in
  environments with diverse geometry and elasticity by promoting
  filopodial-driven adhesion}.
\newblock \emph{\bibinfo{journal}{iScience}} \textbf{\bibinfo{volume}{25}},
  \bibinfo{pages}{104795} (\bibinfo{year}{2022}).

\bibitem{kraus2024masked}
\bibinfo{author}{Kraus, O.} \emph{et~al.}
\newblock \emph{\bibinfo{title}{Masked autoencoders for microscopy are scalable
  learners of cellular biology}}, \bibinfo{pages}{11757--11768}
  (\bibinfo{year}{2024}).

\bibitem{corum}
\bibinfo{author}{Tsitsiridis, G.} \emph{et~al.}
\newblock \bibinfo{title}{Corum: the comprehensive resource of mammalian
  protein complexes–2022}.
\newblock \emph{\bibinfo{journal}{Nucleic Acids Research}}
  \textbf{\bibinfo{volume}{51}}, \bibinfo{pages}{D539--D545}
  (\bibinfo{year}{2022}).
\newblock \urlprefix\url{https://doi.org/10.1093/nar/gkac1015}.

\bibitem{humap}
\bibinfo{author}{Drew, K.} \emph{et~al.}
\newblock \bibinfo{title}{Integration of over 9,000 mass spectrometry
  experiments builds a global map of human protein complexes}.
\newblock \emph{\bibinfo{journal}{Molecular Systems Biology}}
  \textbf{\bibinfo{volume}{13}}, \bibinfo{pages}{932} (\bibinfo{year}{2017}).
\newblock
  \urlprefix\url{https://www.embopress.org/doi/abs/10.15252/msb.20167490}.

\bibitem{reactome}
\bibinfo{author}{Gillespie, M.} \emph{et~al.}
\newblock \bibinfo{title}{The reactome pathway knowledgebase 2022}.
\newblock \emph{\bibinfo{journal}{Nucleic Acids Research}}
  \textbf{\bibinfo{volume}{50}}, \bibinfo{pages}{D687--D692}
  (\bibinfo{year}{2021}).
\newblock \urlprefix\url{https://doi.org/10.1093/nar/gkab1028}.

\bibitem{signor}
\bibinfo{author}{Lo~Surdo, P.} \emph{et~al.}
\newblock \bibinfo{title}{Signor 3.0, the signaling network open resource 3.0:
  2022 update}.
\newblock \emph{\bibinfo{journal}{Nucleic Acids Research}}
  \textbf{\bibinfo{volume}{51}}, \bibinfo{pages}{D631--D637}
  (\bibinfo{year}{2022}).
\newblock \urlprefix\url{https://doi.org/10.1093/nar/gkac883}.

\bibitem{stringdb}
\bibinfo{author}{Szklarczyk, D.} \emph{et~al.}
\newblock \bibinfo{title}{The string database in 2021: customizable
  protein–protein networks, and functional characterization of user-uploaded
  gene/measurement sets}.
\newblock \emph{\bibinfo{journal}{Nucleic Acids Research}}
  \textbf{\bibinfo{volume}{49}}, \bibinfo{pages}{D605--D612}
  (\bibinfo{year}{2020}).
\newblock \urlprefix\url{https://doi.org/10.1093/nar/gkaa1074}.

\bibitem{efaar}
\bibinfo{author}{Celik, S.} \emph{et~al.}
\newblock \bibinfo{title}{Building, benchmarking, and exploring perturbative
  maps of transcriptional and morphological data}.
\newblock \emph{\bibinfo{journal}{PLOS Computational Biology}}
  \textbf{\bibinfo{volume}{20}}, \bibinfo{pages}{1--24} (\bibinfo{year}{2024}).
\newblock \urlprefix\url{https://doi.org/10.1371/journal.pcbi.1012463}.

\bibitem{hagan}
\bibinfo{author}{Sun, L.} \emph{et~al.}
\newblock \bibinfo{title}{Hierarchical amortized gan for 3d high resolution
  medical image synthesis}.
\newblock \emph{\bibinfo{journal}{IEEE Journal of Biomedical and Health
  Informatics}} \textbf{\bibinfo{volume}{26}}, \bibinfo{pages}{3966--3975}
  (\bibinfo{year}{2022}).

\bibitem{fid}
\bibinfo{author}{Heusel, M.}, \bibinfo{author}{Ramsauer, H.},
  \bibinfo{author}{Unterthiner, T.}, \bibinfo{author}{Nessler, B.} \&
  \bibinfo{author}{Hochreiter, S.}
\newblock \bibinfo{title}{Gans trained by a two time-scale update rule converge
  to a local nash equilibrium}.
\newblock \emph{\bibinfo{journal}{Advances in neural information processing
  systems}} \textbf{\bibinfo{volume}{30}} (\bibinfo{year}{2017}).

\bibitem{kynkaanniemi2019improved}
\bibinfo{author}{Kynk{\"a}{\"a}nniemi, T.}, \bibinfo{author}{Karras, T.},
  \bibinfo{author}{Laine, S.}, \bibinfo{author}{Lehtinen, J.} \&
  \bibinfo{author}{Aila, T.}
\newblock \bibinfo{title}{Improved precision and recall metric for assessing
  generative models}.
\newblock \emph{\bibinfo{journal}{Advances in neural information processing
  systems}} \textbf{\bibinfo{volume}{32}} (\bibinfo{year}{2019}).

\bibitem{naeem2020reliable}
\bibinfo{author}{Naeem, M.~F.}, \bibinfo{author}{Oh, S.~J.},
  \bibinfo{author}{Uh, Y.}, \bibinfo{author}{Choi, Y.} \& \bibinfo{author}{Yoo,
  J.}
\newblock \emph{\bibinfo{title}{Reliable fidelity and diversity metrics for
  generative models}}, \bibinfo{pages}{7176--7185}
  (\bibinfo{organization}{PMLR}, \bibinfo{year}{2020}).

\bibitem{zheng2025diffusion}
\bibinfo{author}{Zheng, K.}, \bibinfo{author}{He, G.}, \bibinfo{author}{Chen,
  J.}, \bibinfo{author}{Bao, F.} \& \bibinfo{author}{Zhu, J.}
\newblock \emph{\bibinfo{title}{Diffusion bridge implicit models}}
  (\bibinfo{year}{2025}).
\newblock \urlprefix\url{https://openreview.net/forum?id=eghAocvqBk}.

\bibitem{zhou2024denoising}
\bibinfo{author}{Zhou, L.}, \bibinfo{author}{Lou, A.}, \bibinfo{author}{Khanna,
  S.} \& \bibinfo{author}{Ermon, S.}
\newblock \emph{\bibinfo{title}{Denoising diffusion bridge models}}
  (\bibinfo{year}{2024}).
\newblock \urlprefix\url{https://openreview.net/forum?id=FKksTayvGo}.

\bibitem{Sero2015-ll}
\bibinfo{author}{Sero, J.~E.} \emph{et~al.}
\newblock \bibinfo{title}{Cell shape and the microenvironment regulate nuclear
  translocation of {NF-$\kappa$B} in breast epithelial and tumor cells}.
\newblock \emph{\bibinfo{journal}{Mol Syst Biol}}
  \textbf{\bibinfo{volume}{11}}, \bibinfo{pages}{790} (\bibinfo{year}{2015}).

\bibitem{navidi2025morphodiff}
\bibinfo{author}{Navidi, Z.} \emph{et~al.}
\newblock \emph{\bibinfo{title}{Morphodiff: Cellular morphology painting with
  diffusion models}} (\bibinfo{year}{2025}).
\newblock \urlprefix\url{https://openreview.net/forum?id=PstM8YfhvI}.

\bibitem{Copperman2023}
\bibinfo{author}{Copperman, J.}, \bibinfo{author}{Gross, S.~M.},
  \bibinfo{author}{Chang, Y.~H.}, \bibinfo{author}{Heiser, L.~M.} \&
  \bibinfo{author}{Zuckerman, D.~M.}
\newblock \bibinfo{title}{Morphodynamical cell state description via live-cell
  imaging trajectory embedding} .

\bibitem{Lubba2019}
\bibinfo{author}{Lubba, C.~H.} \emph{et~al.}
\newblock \bibinfo{title}{catch22: Canonical time-series characteristics}.
\newblock \emph{\bibinfo{journal}{Data Mining and Knowledge Discovery}}
  \textbf{\bibinfo{volume}{33}}, \bibinfo{pages}{1821--1852}
  (\bibinfo{year}{2019}).
\newblock \urlprefix\url{https://doi.org/10.1007/s10618-019-00647-x}.

\bibitem{Yin2013}
\bibinfo{author}{Yin, Z.} \emph{et~al.}
\newblock \bibinfo{title}{A screen for morphological complexity identifies
  regulators of switch-like transitions between discrete cell shapes}.
\newblock \emph{\bibinfo{journal}{Nature Cell Biology}}
  \textbf{\bibinfo{volume}{15}}, \bibinfo{pages}{860--871}
  (\bibinfo{year}{2013}).
\newblock \urlprefix\url{https://doi.org/10.1038/ncb2764}.

\bibitem{Cooper2015-jw}
\bibinfo{author}{Cooper, S.}, \bibinfo{author}{Sadok, A.},
  \bibinfo{author}{Bousgouni, V.} \& \bibinfo{author}{Bakal, C.}
\newblock \bibinfo{title}{Apolar and polar transitions drive the conversion
  between amoeboid and mesenchymal shapes in melanoma cells}.
\newblock \emph{\bibinfo{journal}{Mol Biol Cell}}
  \textbf{\bibinfo{volume}{26}}, \bibinfo{pages}{4163--4170}
  (\bibinfo{year}{2015}).

\bibitem{Sailem2015}
\bibinfo{author}{Sailem, H.~Z.}, \bibinfo{author}{Sero, J.~E.} \&
  \bibinfo{author}{Bakal, C.}
\newblock \bibinfo{title}{Visualizing cellular imaging data using phenoplot}.
\newblock \emph{\bibinfo{journal}{Nature Communications}}
  \textbf{\bibinfo{volume}{6}}, \bibinfo{pages}{5825} (\bibinfo{year}{2015}).
\newblock \urlprefix\url{https://doi.org/10.1038/ncomms6825}.

\bibitem{Sero2017}
\bibinfo{author}{Sero, J.~E.} \& \bibinfo{author}{Bakal, C.}
\newblock \bibinfo{title}{Multiparametric analysis of cell shape demonstrates
  that {\&}{\#}x3b2;-pix directly couples yap activation to extracellular
  matrix adhesion}.
\newblock \emph{\bibinfo{journal}{Cell Systems}} \textbf{\bibinfo{volume}{4}},
  \bibinfo{pages}{84--96.e6} (\bibinfo{year}{2017}).
\newblock \urlprefix\url{https://doi.org/10.1016/j.cels.2016.11.015}.

\bibitem{Way2022-bh}
\bibinfo{author}{Way, G.~P.} \emph{et~al.}
\newblock \bibinfo{title}{Morphology and gene expression profiling provide
  complementary information for mapping cell state}.
\newblock \emph{\bibinfo{journal}{Cell Syst}} \textbf{\bibinfo{volume}{13}},
  \bibinfo{pages}{911--923.e9} (\bibinfo{year}{2022}).

\bibitem{10.1098/rsob.130132}
\bibinfo{author}{Sailem, H.}, \bibinfo{author}{Bousgouni, V.},
  \bibinfo{author}{Cooper, S.} \& \bibinfo{author}{Bakal, C.}
\newblock \bibinfo{title}{Cross-talk between rho and rac gtpases drives
  deterministic exploration of cellular shape space and morphological
  heterogeneity}.
\newblock \emph{\bibinfo{journal}{Open Biology}} \textbf{\bibinfo{volume}{4}},
  \bibinfo{pages}{130132} (\bibinfo{year}{2014}).

\bibitem{Way2022-eo}
\bibinfo{author}{Way, G.~P.} \emph{et~al.}
\newblock \bibinfo{title}{Morphology and gene expression profiling provide
  complementary information for mapping cell state}.
\newblock \emph{\bibinfo{journal}{Cell Syst}} \textbf{\bibinfo{volume}{13}},
  \bibinfo{pages}{911--923.e9} (\bibinfo{year}{2022}).

\bibitem{Simpson306571}
\bibinfo{author}{Simpson, C.~M.}, \bibinfo{author}{Ferrari, N.},
  \bibinfo{author}{Calvo, F.} \& \bibinfo{author}{Bakal, C.}
\newblock \bibinfo{title}{The dynamics of erk signaling in melanoma, and the
  response to braf or mek inhibition, are cell cycle dependent}.
\newblock \emph{\bibinfo{journal}{bioRxiv}}  (\bibinfo{year}{2018}).
\newblock
  \urlprefix\url{https://www.biorxiv.org/content/early/2018/08/13/306571}.

\bibitem{Kudo2018}
\bibinfo{author}{Kudo, T.} \emph{et~al.}
\newblock \bibinfo{title}{Live-cell measurements of kinase activity in single
  cells using translocation reporters}.
\newblock \emph{\bibinfo{journal}{Nature Protocols}}
  \textbf{\bibinfo{volume}{13}}, \bibinfo{pages}{155--169}
  (\bibinfo{year}{2018}).
\newblock \urlprefix\url{https://doi.org/10.1038/nprot.2017.128}.

\bibitem{pascual2021multiplexed}
\bibinfo{author}{Pascual-Vargas, P.}, \bibinfo{author}{Arias-Garcia, M.},
  \bibinfo{author}{Roumeliotis, T.}, \bibinfo{author}{Choudhary, J.~S.} \&
  \bibinfo{author}{Bakal, C.}
\newblock \bibinfo{title}{Multiplexed quantitative screens of single cell shape
  and yap/taz localisation identify dock5 as a coincident detector of polarity
  and adhesion during migration}.
\newblock \emph{\bibinfo{journal}{TAZ Localisation Identify DOCK5 as a
  Coincident Detector of Polarity and Adhesion During Migration}}
  (\bibinfo{year}{2021}).

\bibitem{10.1145/37401.37422}
\bibinfo{author}{Lorensen, W.~E.} \& \bibinfo{author}{Cline, H.~E.}
\newblock \emph{\bibinfo{title}{Marching cubes: A high resolution 3d surface
  construction algorithm}}, SIGGRAPH '87, \bibinfo{pages}{163–169}
  (\bibinfo{publisher}{Association for Computing Machinery},
  \bibinfo{address}{New York, NY, USA}, \bibinfo{year}{1987}).
\newblock \urlprefix\url{https://doi.org/10.1145/37401.37422}.

\bibitem{Lewiner01012003}
\bibinfo{author}{Lewiner, T.}, \bibinfo{author}{Lopes, H.},
  \bibinfo{author}{Vieira, A.~W.} \& \bibinfo{author}{and, G.~T.}
\newblock \bibinfo{title}{Efficient implementation of marching cubes' cases
  with topological guarantees}.
\newblock \emph{\bibinfo{journal}{Journal of Graphics Tools}}
  \textbf{\bibinfo{volume}{8}}, \bibinfo{pages}{1--15} (\bibinfo{year}{2003}).
\newblock \urlprefix\url{https://doi.org/10.1080/10867651.2003.10487582}.

\bibitem{10.5555/1953048.2078195}
\bibinfo{author}{Pedregosa, F.} \emph{et~al.}
\newblock \bibinfo{title}{Scikit-learn: Machine learning in python}.
\newblock \emph{\bibinfo{journal}{J. Mach. Learn. Res.}}
  \textbf{\bibinfo{volume}{12}}, \bibinfo{pages}{2825–2830}
  (\bibinfo{year}{2011}).

\bibitem{buitinck2013apidesignmachinelearning}
\bibinfo{author}{Buitinck, L.} \emph{et~al.}
\newblock \bibinfo{title}{Api design for machine learning software: experiences
  from the scikit-learn project} (\bibinfo{year}{2013}).
\newblock \urlprefix\url{https://arxiv.org/abs/1309.0238}.
\newblock
  \bibinfo{eprint}{{\href{https://arxiv.org/abs/1309.0238}{{arXiv:1309.0238}}}}.

\bibitem{Holmes2015}
\bibinfo{author}{Holmes, A.~J.} \emph{et~al.}
\newblock \bibinfo{title}{Brain genomics superstruct project initial data
  release with structural, functional, and behavioral measures}.
\newblock \emph{\bibinfo{journal}{Scientific Data}}
  \textbf{\bibinfo{volume}{2}}, \bibinfo{pages}{150031} (\bibinfo{year}{2015}).
\newblock \urlprefix\url{https://doi.org/10.1038/sdata.2015.31}.

\bibitem{Regan2010-ul}
\bibinfo{author}{Regan, E.~A.} \emph{et~al.}
\newblock \bibinfo{title}{Genetic epidemiology of {COPD} ({COPDGene}) study
  design}.
\newblock \emph{\bibinfo{journal}{COPD}} \textbf{\bibinfo{volume}{7}},
  \bibinfo{pages}{32--43} (\bibinfo{year}{2010}).

\bibitem{Petersen2009-bh}
\bibinfo{author}{Petersen, R.~C.} \emph{et~al.}
\newblock \bibinfo{title}{Alzheimer's disease neuroimaging initiative ({ADNI)}:
  clinical characterization}.
\newblock \emph{\bibinfo{journal}{Neurology}} \textbf{\bibinfo{volume}{74}},
  \bibinfo{pages}{201--209} (\bibinfo{year}{2009}).

\bibitem{Armato2015}
\bibinfo{author}{Armato~III, S.~G.} \emph{et~al.}
\newblock \bibinfo{title}{Data from lidc-idri}.
\newblock \bibinfo{howpublished}{The Cancer Imaging Archive}
  (\bibinfo{year}{2015}).
\newblock \urlprefix\url{https://doi.org/10.7937/K9/TCIA.2015.LO9QL9SX}.

\bibitem{Saha2021}
\bibinfo{author}{Saha, A.} \emph{et~al.}
\newblock \bibinfo{title}{Dynamic contrast-enhanced magnetic resonance images
  of breast cancer patients with tumor locations [data set]}.
\newblock \bibinfo{howpublished}{The Cancer Imaging Archive}
  (\bibinfo{year}{2021}).
\newblock \urlprefix\url{https://doi.org/10.7937/TCIA.e3sv-re93}.

\bibitem{10.1371/journal.pmed.1002699}
\bibinfo{author}{Bien, N.} \emph{et~al.}
\newblock \bibinfo{title}{Deep-learning-assisted diagnosis for knee magnetic
  resonance imaging: Development and retrospective validation of mrnet}.
\newblock \emph{\bibinfo{journal}{PLOS Medicine}}
  \textbf{\bibinfo{volume}{15}}, \bibinfo{pages}{1--19} (\bibinfo{year}{2018}).
\newblock \urlprefix\url{https://doi.org/10.1371/journal.pmed.1002699}.

\bibitem{7780677}
\bibinfo{author}{Szegedy, C.}, \bibinfo{author}{Vanhoucke, V.},
  \bibinfo{author}{Ioffe, S.}, \bibinfo{author}{Shlens, J.} \&
  \bibinfo{author}{Wojna, Z.}
\newblock \emph{\bibinfo{title}{Rethinking the inception architecture for
  computer vision}}, \bibinfo{pages}{2818--2826} (\bibinfo{year}{2016}).

\bibitem{5206848}
\bibinfo{author}{Deng, J.} \emph{et~al.}
\newblock \emph{\bibinfo{title}{Imagenet: A large-scale hierarchical image
  database}}, \bibinfo{pages}{248--255} (\bibinfo{year}{2009}).

\end{thebibliography}

\newpage

\section{Supplementary Materials}

\begin{appendices}

\begin{table}[h]
\centering
\resizebox{\textwidth}{!}{%
\begin{tabular}{lccccccccc}
\toprule
\textbf{Method} & \multicolumn{3}{c}{$FID^{-1}$ (↑)} & \multicolumn{3}{c}{F1 Score (↑)} & \multicolumn{3}{c}{Coverage (↑)} \\
 & Binimetinib & Blebbistatin & Nocodazole & Binimetinib & Blebbistatin & Nocodazole & Binimetinib & Blebbistatin & Nocodazole \\
\midrule
\textsc{Form} & 0.830 & 0.635 & 1.000 & 0.624 & 0.458 & 0.629 & 0.747 & 0.626 & 0.849 \\
HA-GAN \citep{hagan} & 0.000 & 0.004 & 0.022 & 0.173 & 0.231 & 0.153 & 0.596 & 0.560 & 0.798 \\
MedicalDiffusion \citep{khader2023denoising} & 0.025 & 0.032 & 0.061 & 0.138 & 0.285 & 0.122 & 0.599 & 0.598 & 0.808 \\
\bottomrule
\end{tabular}%
}
\caption{Comparison of generative models across three metrics: FID, F1 score, and coverage for each perturbation setting. \textsc{Form} outperforms baseline methods across all metrics in each perturbation setting.}
\label{tab:model_comparison}
\end{table}

\begin{figure}[htbp]
    \centering
    \includegraphics[width=\linewidth]{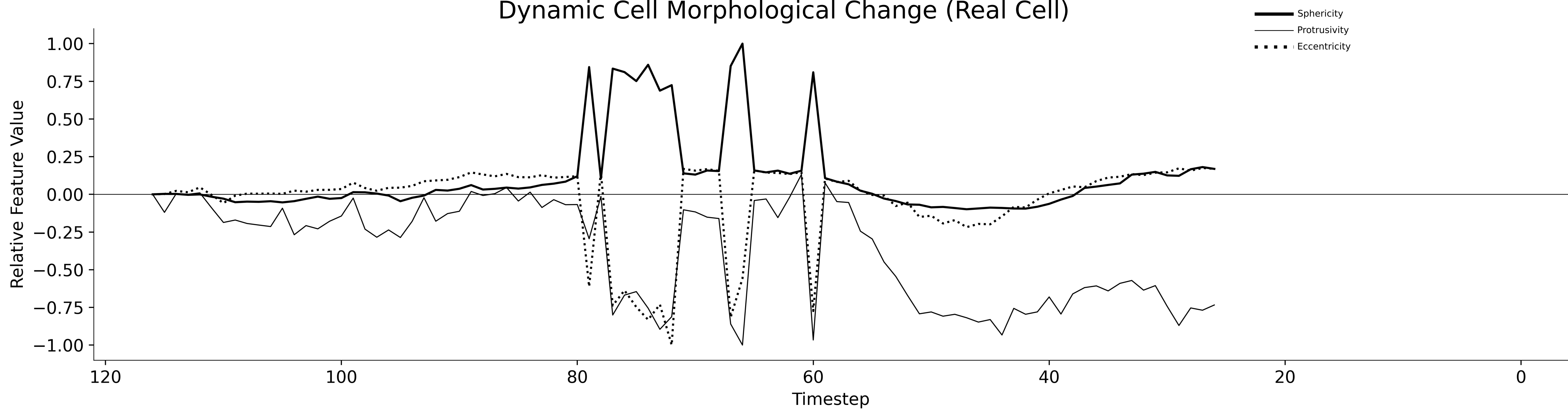}
    \caption{\textbf{Dynamic morphology of a live cell under treatment.} 
    Time-lapse imaging of a single cell at five-minute intervals over a ten-hour period reveals that morphological change proceeds through abrupt, heterogeneous shifts rather than smooth, linear transitions. }
 
    \label{supp2}
\end{figure}

\begin{figure}[htbp]
    \centering
    \includegraphics[width=\linewidth]{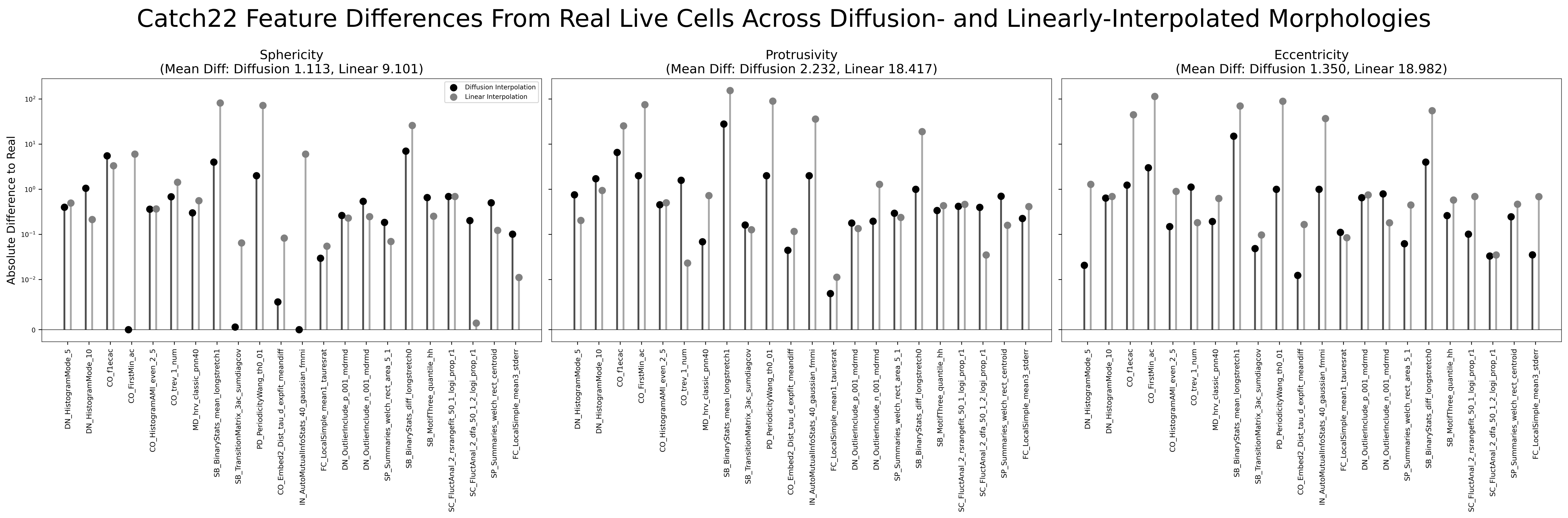}
    \caption{\textbf{Catch22 feature comparison of real and generated morphodynamics.}  
    Absolute differences in Catch22 time-series features between real live-cell dynamics, \textsc{Form}-generated trajectories, and linear interpolations show that \textsc{Form} more closely recapitulates true morphological evolution than interpolation-based approaches.}
    \label{supp3}
\end{figure}

\begin{table}[h]
    \centering
   
    \label{tab:vqgan_hyperparams}
    \begin{tabular}{ll}
        \toprule
        \textbf{Hyperparameter} & \textbf{Value} \\
        \midrule
        \textbf{Learning Rate} & $3 \times 10^{-4}$ \\
        \textbf{Batch Size} & 2 \\
        \textbf{Latent Dimension (per channel)} & 16 \\
        \textbf{Training Steps} & 100,000 \\
        \textbf{Codebook Size (per codebook)} & 1024 \\
        \textbf{Reconstruction Loss} & Mean Squared Error (MSE) \\
        \textbf{Commitment Loss Weight} & 0.25 \\
        \textbf{Optimizer} & Adam \\
        \textbf{Beta 1 (Adam)} & 0.9 \\
        \textbf{Beta 2 (Adam)} & 0.99 \\
        \bottomrule
    \end{tabular}
     \caption{VQGAN Hyperparameters}
\end{table}

\begin{table}[h]
    \centering
    
    \label{tab:DualChannelUNet_hyperparams}
    \begin{tabular}{ll}
        \toprule
        \textbf{Hyperparameter} & \textbf{Value} \\
        \midrule
        \textbf{Learning Rate} & $1 \times 10^{-4}$ \\
        \textbf{Batch Size} & 2 \\
        \textbf{Number of Timesteps} & 1000 \\
        \textbf{Loss Function} & L1 Loss \\
        \textbf{Number of Channels} & 2 (Cell, Nucleus) \\
        \textbf{3D Convolution Kernel Size} & $3 \times 3 \times 3$ \\
        \textbf{Dimension Multiplier} & [1,2,4,8] \\
        \textbf{Number of Attention Layers} & 2 (Spatial and Depth-wise) \\
        \textbf{Optimizer} & Adam \\
        \textbf{Beta 1 (Adam)} & 0.9 \\
        \textbf{Beta 2 (Adam)} & 0.99 \\
        \textbf{Normalisation} & Instance Normalisation \\
        \textbf{ema decay} & 0.995 \\
        \bottomrule
    \end{tabular}
    \caption{DDPM Hyperparameters}
\end{table}




\end{appendices}

\end{document}